\documentclass[10pt,twocolumn,letterpaper]{article}

\usepackage[pagenumbers]{cvpr} 

\usepackage[dvipsnames]{xcolor}

\definecolor{cvprblue}{rgb}{0.21,0.49,0.74}
\usepackage[pagebackref,breaklinks,colorlinks,citecolor=cvprblue]{hyperref}
\usepackage{amsmath}
\usepackage{amssymb}
\usepackage{booktabs}

\usepackage{url}            
\usepackage{amsfonts}       
\usepackage{microtype}      
\usepackage{xcolor}         
\usepackage{multirow}
\usepackage{colortbl}
\usepackage{enumitem}
\usepackage{subcaption}
\usepackage{newtxmath}
\usepackage{tcolorbox}
\tcbuselibrary{skins,breakable}
\usepackage[accsupp]{axessibility}

\usepackage{scalerel}  
\def\logo{\scalerel*{\includegraphics{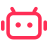}}{\textrm{\textbigcircle}}}
\def\shikralogo{\scalerel*{\includegraphics{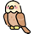}}{\textrm{\textbigcircle}}}
\def\sad{\scalerel*{\includegraphics{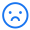}}{\textrm{\textbigcircle}}}
\def\qwen{\scalerel*{\includegraphics{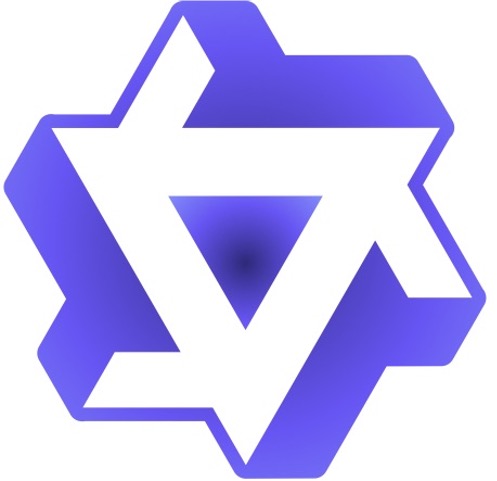}}{\textrm{\textbigcircle}}}

\def\confName{CVPR}
\def\confYear{2024}

\title{Pink: Unveiling \textbf{\textcolor{pink}{the}} Po\textbf{\textcolor{pink}{w}}er of Referential Comprehension for Multi-mod\textbf{\textcolor{pink}{a}}l \textbf{\textcolor{pink}{LL}}Ms}

\author{Shiyu~Xuan$^{1}$\thanks{This work was done during the internship of the first author at Ant Group.}, Qingpei~Guo$^{2}$, Ming~Yang$^{2}$, Shiliang~Zhang$^{1}$\\
  $^{1}$National Key Laboratory for Multimedia Information Processing, \\
  School of Computer Science, Peking University, Beijing, China. \\
  $^{2}$Ant Group \\
  {\tt\small shiyu\_xuan@stu.pku.edu.cn, \{qingpei.gqp, m.yang\}@antgroup.com, slzhang.jdl@pku.edu.cn}
}

\begin{document}
\maketitle
\begin{abstract}
    Multi-modal Large Language Models (MLLMs) have shown remarkable capabilities in various multi-modal tasks. Nevertheless, their performance in fine-grained image understanding tasks is still limited.
    To address this issue, this paper proposes a new framework to enhance the fine-grained image understanding abilities of MLLMs.
    Specifically, we present a new method for constructing the instruction tuning dataset at a low cost by leveraging annotations in existing datasets.
    A self-consistent bootstrapping method is also introduced to extend existing dense object annotations into high-quality referring-expression-bounding-box pairs.
    These methods enable the generation of high-quality instruction data which includes a wide range of fundamental abilities essential for fine-grained image perception.
    Moreover, we argue that the visual encoder should be tuned during instruction tuning to mitigate the gap between full image perception and fine-grained image perception.
    Experimental results demonstrate the superior performance of our method. For instance, our model exhibits a 5.2\% accuracy improvement over Qwen-VL on GQA and surpasses the accuracy of Kosmos-2 by 24.7\% on RefCOCO\_val. We have also attained the top rank on the leaderboard of MMBench. This promising performance is achieved by training on only publicly available data, making it easily reproducible.
    The models, datasets, and codes are publicly available at~\url{https://github.com/SY-Xuan/Pink}.
\end{abstract}
    
\section{Introduction}
\label{sec:intro}

Large Language Models (LLMs)~\cite{gpt3,t5,llama,bloom} show impressive capabilities across a wide range of natural language tasks.
These inspiring results have motivated researchers to extend LLMs to Multi-modal Large Language Models (MLLMs) by integrating additional modalities, \emph{e.g.}, image, audio, or point cloud.
Visual instruction tuning~\cite{llava,dai2023instructblip,ye2023mplug}, using high-quality image-text instruction tuning data, allows the incorporation of visual comprehension ability into LLMs by projecting visual features into the natural language space of the LLMs~\cite{yin2023survey}.
Powered by those methods, existing MLLMs are capable of basic image-level comprehension. However, they are still confronted by fine-grained image understanding~\cite{wang2023contextual,wang2023humvis,kazemzadeh2014referitgame}.
Limited fine-grained image understanding ability hinders the performance of MLLMs in multi-modal tasks and restricts their potential applications as reported in the GPT-4V(ision) test report~\cite{yang2023dawn}.

\begin{figure}
    \begin{center}
    \includegraphics[width=0.61\linewidth]{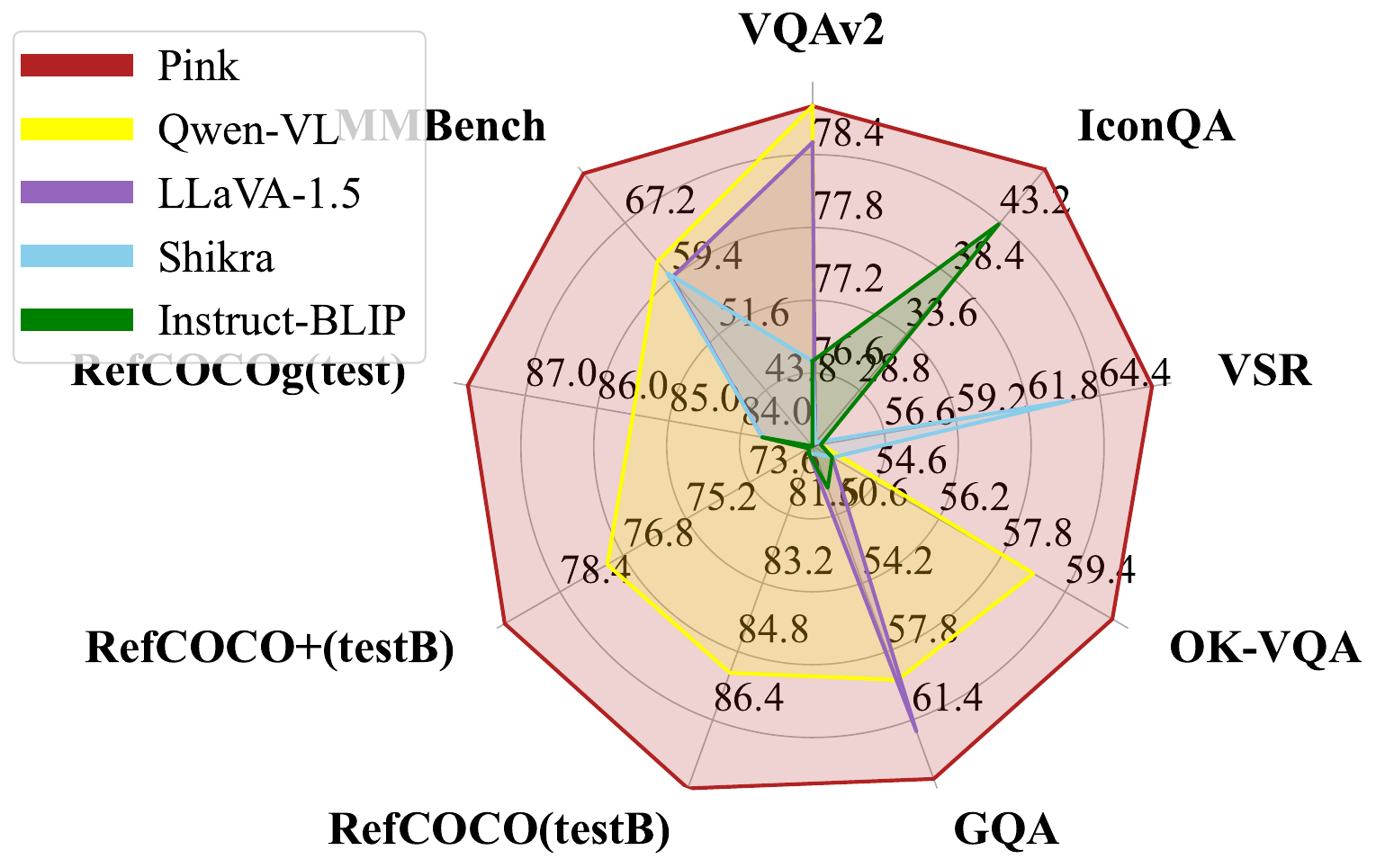}
    \includegraphics[width=0.38\linewidth]{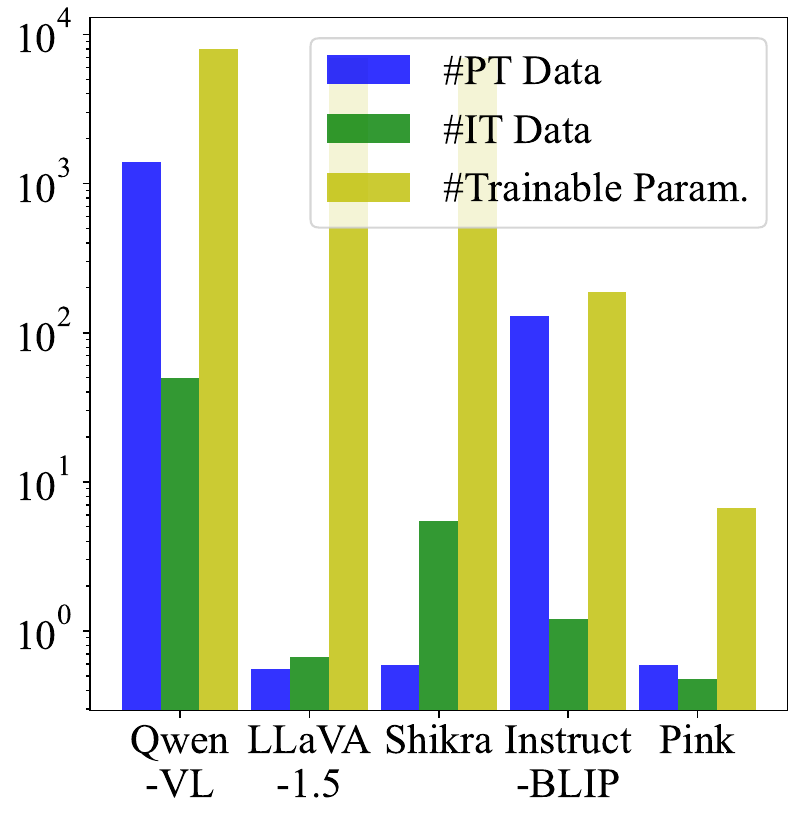}
    \caption{With fewer trainable parameters and less training data, Pink achieves the best performance on both conventional multi-modal tasks and RC tasks. ``\#Trainable Param.'', ``\#PT Data'', and ``\#IT Data'' indicate the number of trainable parameters, the number of samples in pre-training and instruction tuning stage, respectively.}
    \label{fig:motivation}
    \end{center}
\end{figure}

To address this issue, some methods~\cite{chen2023shikra,peng2023kosmos} incorporate some datasets related to Referential Comprehension (RC) such as RefCOCO~\cite{kazemzadeh2014referitgame}, and PointQA~\cite{mani2020point} to enhance the fine-grained image perception ability of MLLMs. However, these datasets are insufficient to cover a wide range of abilities that MLLMs desire to have for fine-grained image perception. Limited RC tasks also make it hard for the model to generalize across various RC tasks through instruction tuning. For instance, as shown in Fig.~\ref{fig:vis_main}, Shikra~\cite{chen2023shikra} and Qwen-VL~\cite{bai2023qwen} show limited instruction-following ability on RC tasks beyond its instruction tuning, failing to provide relevant responses to questions.

In addition to instruction tuning, the capability of the visual encoder is also important to the performance of MLLMs.
MLLMs typically employ a visual encoder trained through contrastive language-image pre-training like the one in CLIP~\cite{radford2021learning}.
Simply performing global alignment is ineffective in exploring fine-grained relationships between image regions and text words~\cite{zhong2022regionclip,wu2023clipself}. The visual encoder stands as a bottleneck for achieving fine-grained image perception in MLLMs.

This paper proposes a new framework to enhance the fine-grained image perception ability of MLLMs through RC tasks. We refer to the trained model as \textbf{Pink} \logo \footnote{This name is from the main character of the album \textit{The Wall} by the great rock band \textit{Pink Floyd}.}.
Fine-grained image understanding is closely tied to some fundamental abilities such as instance identification and recognition of relative positions between different instances. Integrating tasks that demand these fundamental abilities during instruction tuning is crucial for enhancing the model's fine-grained image perception ability.
To this end, we propose a new dataset construction pipeline that extends the annotations of existing datasets to various RC tasks about these fundamental abilities. Specifically, we design several RC tasks, such as visual relation reasoning and visual spatial reasoning, based on the annotations from Visual Genome~\cite{krishna2017visual}.
To further incorporate more training data for these RC tasks, we introduce a novel self-consistent bootstrapping method to extend dense object annotations to referring-expression-bounding-box pairs.
Compared to the expensive, and uncontrollable process of generating data using the GPT4 API~\cite{chen2023shikra,zhao2023chatspot}, our method leverages the existing annotations from datasets. This approach results in high-quality data and precise enhancement of the necessary capabilities required by the model. As shown in Fig.~\ref{fig:motivation}, the high-quality instruction tuning data generated by our method enables the model to achieve promising performance with a reduced number of training samples.

Improving the fine-grained image understanding ability of the visual encoder is not a trivial task. Most existing MLLMs~\cite{minigpt4,llava,ye2023mplug,chen2023shikra,dai2023instructblip} freeze the visual encoder during instruction tuning. Because directly tuning the visual encoder can result in a semantic loss due to the limited scale of the visual instruction dataset~\cite{wang2023makes}. To address this issue, we tune the visual encoder by introducing several tunable components like Adapters~\cite{houlsby2019parameter} and LoRA~\cite{lora}. Freezing the main parameters of the model avoids forgetting the learned knowledge. The introduced tunable components are trained to adapt the visual encoder.

We have conducted extensive experiments to test the performance of the model. Benefited by the designed tasks in the instruction tuning, our framework enhances MLLMs' performance in both conventional vision-language tasks and RC tasks. For instance, with only 6.7M tunable parameters, we achieve up to a 6.0\% accuracy improvement on OK-VQA~\cite{marino2019ok} compared to Shikra~\cite{chen2023shikra}. We also attain the top rank on the leaderboard of MMBench~\cite{liu2023mmbench}. It should be noted that, our method also surpasses methods that rely on more training data, \emph{e.g.}, Qwen-VL~\cite{bai2023qwen}.

This work is an original effort to enhance MLLMs' fine-grained image perception ability by addressing two main bottlenecks: limited instruction tuning tasks and the lack of ability of the visual encoder.
By designing tasks related to fundamental abilities, every dataset with corresponding annotations can be converted into the instruction tuning dataset. The self-consistent bootstrapping further increases the number of training data. This dataset construction pipeline significantly reduces the cost of obtaining high-quality data with diversified tasks and eliminates the dependency on GPT4 APIs.
The whole training pipeline is reproducible in academia as it only relies on publicly available data and can be trained on consumer GPUs with 24GB memory.
We will release the codes and datasets to facilitate further research and evaluation.

\section{Related Works}
\label{sec:related work}

\noindent\textbf{Multi-modal large language model.}
Several approaches have been proposed to condition LLMs with additional modalities. Typically, these methods utilize two-stage training. The pre-training stage is performed to align two modalities with image-text pairs. The subsequent stage is adopted to improve the ability of MLLMs to follow instructions with high-quality instruction tuning dataset~\cite{llava,minigpt4,baize,liu2023aligning}. Many methods freeze the visual encoder during the pre-training stage to reduce requirement of large-scale image-text pairs. For example, Mini-GPT4~\cite{minigpt4} and LLaVA~\cite{llava} only fine-tune a single fully connected layer to align the vision and language modalities.
Other methods leverage millions of image-text pairs to achieve better alignment between two modalities.
Instruct-BLIP~\cite{dai2023instructblip} introduces an instruction-aware visual feature extraction method and fine-tunes the entire Q-Former, showing promising zero-shot generalization ability on various multi-modal tasks.
mPlug-Owl~\cite{ye2023mplug} incorporates a visual abstractor module to align the two-modalities. Both the visual encoder and the visual abstractor are updated during the pre-training stage. All of above methods freeze the visual encoder during the multi-modal instruction tuning stage to prevent the potential semantic loss caused by the small-scale instruction tuning dataset. However, this strategy makes the visual encoder cannot benefit from the multi-modal instruction tuning.

\noindent\textbf{Referential Comprehension of MLLMs.}
Referential comprehension is important to the fine-grained image perception of MLLMs. Therefore, enhancing MLLMs with the RC ability is highly valuable. Inspired by Pix2Seq~\cite{chen2021pix2seq}, many works use discrete coordinate tokens to encode spatial information and unify RC tasks as sequence generation tasks, \emph{e.g.}, OFA~\cite{wang2022ofa}, Unified-io~\cite{lu2022unified}, and Kosmos-2~\cite{peng2023kosmos}. Another line of works, as seen in PVIT~\cite{chen2023positionenhanced} and GPT4RoI~\cite{zhang2023gpt4roi}, leverage the ROI operation~\cite{he2017mask} to extract features of referring objects. These works require extra modules and may lose context information. Another limitation of these works is that they cannot refer objects in their responses, limiting their applications, \emph{e.g.}, in visual grounding.

In addition to the model design, the construction of RC instruction tuning data also plays a crucial role. Shikra~\cite{chen2023shikra} converts existing datasets of RC tasks including RefCOCO~\cite{kazemzadeh2014referitgame} and PointQA~\cite{mani2020point} into the instruction following format. Kosmos-2 uses the grounding model GLIP~\cite{li2022grounded} to extract coordinates of noun chunks in image captions and constructs a large-scale dataset. Datasets constructed by the above methods only includes RC tasks, such as visual grounding, grounding caption and pointQA, which are still not diversified enough to cover various RC tasks. The trained models thus show a poor ability to generalize to new RC tasks beyond the instruction tuning. ChatSpot~\cite{zhao2023chatspot}, PVIT~\cite{chen2023positionenhanced}, and Shikra~\cite{chen2023shikra} all prompt GPT4 to generate instruction tuning data for RC, which is expensive and uncontrollable.

\noindent\textbf{Differences with previous works.}
Existing methods for enhancing MLLMs through RC tasks construct the instruction tuning dataset by either integrating existing RC datasets or relying on GPT4 APIs.
However, these methods exhibit some major drawbacks: 1) the diversity of RC tasks cannot cover a wide range of fundamental abilities, and 2) data generation through GPT4 APIs is expensive, uncontrollable, and prone to noise.
In contrast, our work effectively leverages existing datasets to cover a wide variety of RC tasks. The proposed self-consistent bootstrapping method extends dense object annotations to referring-expression-bounding-box pairs. This pipeline significantly reduces the cost of generating high-quality datasets.
The high qualify of the data allows our model to be trained with fewer parameters on less training data, which is friendly to reproduce in academia, than large commercial MLLMs.

\section{Methodology}
\subsection{Model Architecture and Training Pipeline}
\begin{figure}
    \begin{center}
    \includegraphics[width=1.0\linewidth]{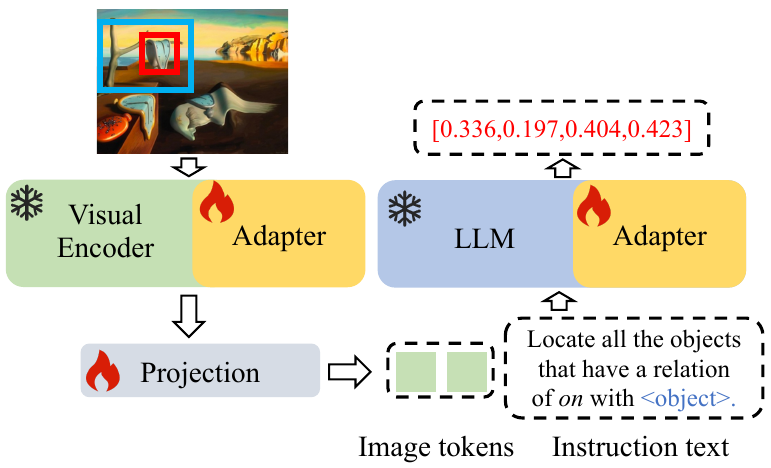}
    \caption{The illustration of our Pink model. Pink follows the architecture of LLaVA~\cite{llava}, which consists of three main components: a visual encoder, a projection layer, and a decoder-only LLM. The coordinates of a bounding box are converted into texts in a specific format. During instruction tuning, we freeze the visual encoder and LLM and only update the Adapters and the projection layer. 
    }
    \label{fig:framework}
    \end{center}
\end{figure}

\noindent\textbf{Model architecture.}
As shown in Fig.~\ref{fig:framework}, Pink follows a similar architecture of LLaVA~\cite{llava}, which consists of a visual encoder $\Phi_{V}$, a projection layer $\Phi_{P}$, and a decoder-only LLM $\Phi_{L}$. Given an image $I$ and a sequence of word embeddings $Q_T$ representing an instruction sentence, the visual encoder is employed to embed the image as a sequence of visual tokens $Z_V = \Phi_V(I)$. A linear layer is used as $\Phi_{P}$ to convert $Z_V$ into the input space of the LLM $Z_T = \Phi_{P}(Z_V)$. $Z_T$ and $Q_T$ are concatenated and fed into $\Phi_{L}$ to generate the next word.

To enable the LLMs to take coordinates as input and output, similar to Shikra~\cite{chen2023shikra}, the coordinates are converted into texts in a specific format. Specifically, for a bounding box represented by its coordinates of the top-left and bottom-right corners $[x_{min}, y_{min}, x_{max}, y_{max}]$, the coordinates are normalized to the range $[0,1]$ with respect to the image size and retain 3 decimal places for each number, \emph{e.g.}, \textcolor{red}{[0.222,0.333,0.444,0.555]}. This design allows the coordinates to be processed as regular text and can appear in both the input and output of the model.

The visual encoder pre-trained by the contrastive loss~\cite{radford2021learning} lacks region-level image comprehension.
Directly fine-tuning the entire visual encoder during instruction tuning could lead a semantic loss due to the limited instruction tuning data~\cite{wang2023makes}.
To incorporate the fine-grained image perception ability of the visual encoder through the multi-modal instruction tuning with RC tasks, we freeze the visual encoder, meanwhile introducing tunable modules into it.
This approach prevents the visual encoder from suffering semantic loss and provides an efficient way to adapt the model.
In particular, we employ the Adapter~\cite{houlsby2019parameter} at both the visual encoder and LLM. Given an input token feature $Z \in \mathbb{R}^d$, the architecture of an Adapter is defined as follows,
\begin{equation}
    \hat{Z} = \sigma \left(Z W_{d}\right)W_{u} + Z,
    \label{eq:adapter}
\end{equation}
where $W_{d} \in \mathbb{R}^{d \times d_s}$ and $W_{u} \in \mathbb{R}^{d_s \times d}$ denote the weight matrices, $d_s$ is the hidden dimension which is much smaller than $d$, and $\sigma$ denotes the non-linear activation function. $W_{u}$ is initialized to zero to ensure that at the beginning of the training, the Adapter does not change the original output.

\noindent\textbf{Training pipeline.} Both the image and coordinates are mapped into the input space of the LLM. Consequently, the model can be trained end-to-end using a language modeling task, which predicts the next word token based on the preceding context.

The model is trained in two stages.
In the first stage, we exclusively fine-tune the projection layer with a small set of image-text pairs. In the second stage, we freeze both the visual encoder and LLM and fine-tune the newly added Adapters and the projection layer with the instruction tuning dataset. Therefore, both the visual and language modalities can benefit from the multi-modal instruction tuning.

\subsection{Instruction tuning Dataset Construction}
To create the instruction tuning dataset, we unify all the multi-modal tasks into a vision-language dialogue format:
\begin{tcolorbox}
    \textbf{Image:} \{Image tokens\} \\
    \textbf{User:} \{Instruction template\} \\
    \textbf{Assistant:} \{Response\}
\end{tcolorbox}
\noindent where the placeholders \{Image tokens\}, \{Instruction template\}, and \{Response\} will be replaced with the image tokens extracted by $\Phi_{V}$, task instruction template, and the response, respectively.

It is important to introduce diverse RC tasks for the instruction tuning to cover a wide range of fundamental abilities for the find-grained image perception of MLLMs.
Existing datasets only offer limited RC tasks, \emph{e.g.}, visual grounding, grounding caption~\cite{kazemzadeh2014referitgame}, and pointQA~\cite{mani2020point}. Besides the RC tasks mentioned above, we design more diversified RC tasks by incorporating annotations from Visual Genome~\cite{krishna2017visual}, which contain information about region descriptions, objects, and relations between different objects. Following parts proceed to introduce those tasks.

\noindent\textbf{Visual relation reasoning.}
Visual Genome has annotated millions of relationship triplets (\emph{subject-predicate-object}), \emph{e.g.}, man-wearing-hat. We design two types of visual relation reasoning tasks by leveraging these annotations to help the model understand visual relationships between different objects: (1) We randomly select a relationship triplet. Given the coordinates of \emph{subject} and \emph{object}, the model is required to predict their relation. (2) We randomly select one \emph{subject} and a relation from the annotations. The model is required to detect all objects that have selected relation with \emph{subject} and output their coordinates and class names.

\noindent\textbf{Coarse visual spatial reasoning.}
We introduce a coarse visual spatial reasoning task by utilizing the object annotations from Visual Genome. This task enhances MLLMs to identify relative spatial relation between different instances. We define four coarse spatial positions as \emph{top-left, top-right, bottom-left, and bottom-right}. Given a randomly selected object and a coarse spatial position, the model is required to identify all objects located at this position relative to the selected object and predict their coordinates and class names.

\noindent\textbf{Object counting.}
To endow the model with the concept of different instances and the capability of fine-grained object recognition, we design an object counting task. This task requires the model to count objects in the image that belong to the same category as the given object or class name.

\noindent\textbf{Object detection.}
Object detection can empower the model to locate the position and boundaries of objects. This task is also important for the model to identify the existence or category of a certain object.
Given a class name or a selected object, the model is asked to identify all objects that belong to the same category of the given object or class name, and provide their coordinates.

By incorporating these RC tasks into the instruction tuning, the model can learn a variety of RC abilities. To clarify the designed tasks, we list some instruction templates as follows,
\begin{tcolorbox}
    \textbf{Visual Relation Reasoning:} \\
    \textbf{User:} Assist me in finding the relation between \textless subject\textgreater~and \textless object\textgreater~in the photo. \\
    \textbf{Assistant:} \textless relation\textgreater. \\ \\
    \textbf{User:} Please locate and categorize all the objects that have a relation of \textless relation\textgreater~with \textless subject\textgreater. \\
    \textbf{Assistant:} \textless object\textgreater~\textless category\textgreater~\textless object\textgreater~\textless category\textgreater. \\ \\
    \textbf{Coarse Visual Spatial Reasoning:} \\
    \textbf{User:} Identify the objects located at \textless loc\textgreater~of \textless object\textgreater. \\
    \textbf{Assistant:} \textless object\textgreater~\textless category\textgreater~\textless object\textgreater~\textless category\textgreater. \\ \\
    \textbf{Object Counting:} \\
    \textbf{User:} How many objects in the image are of the same category as \textless object\textgreater. \\
    \textbf{Assistant:} \textless number\textgreater. \\ \\
    \textbf{Object Detection:} \\
    \textbf{User:} Identify all the objects that fit the same category as \textless object\textgreater~and display their coordinates. \\
    \textbf{Assistant:} \textless object\textgreater~\textless object\textgreater.
\end{tcolorbox}
\noindent where the placeholders \textless object\textgreater~and \textless category\textgreater~will be replaced with the bounding box coordinates and the class name of a referring object, respectively. \textless subject\textgreater~will be replaced with the bounding box coordinates of the selected subject. \textless relation\textgreater, \textless loc\textgreater, and \textless number\textgreater~will be replaced with the relation between different objects, the selected relative spatial position, and the number of the objects, respectively.
All instruction templates can be found in Supplementary Material.

\begin{figure}
    \begin{center}
    \includegraphics[width=1.0\linewidth]{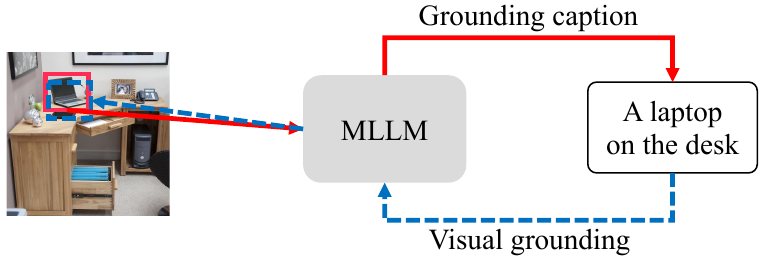}
    \caption{The illustration of self-consistent bootstrapping method. Given a bounding box, our method first generates its description by asking MLLM to perform grounding caption, then leverages the visual grounding to locate the generated description. The low-quality description will be filtered if the IOU between the predicted and ground-truth bounding box is below a threshold.}
    \label{fig:self}
    \end{center}
\end{figure}

\subsection{Self-consistent Bootstrapping Method}
The constructed instruction-following datasets are adopted to reinforce the fine-grained image perception ability of MLLMs. We further acquire more high quality data.
Existing datasets for object detection provide valuable bounding box annotations for objects appearing in the image, making them promising resources for instruction tuning. We propose a self-consistent bootstrapping method by leveraging those datasets. This method extends bounding box annotations to the referring-expression-bounding-box pairs. It comprises two key stages: bounding box description bootstrapping and self-consistent filtering as shown in Fig.~\ref{fig:self}.

At the bounding box description bootstrapping stage, given a bounding box $B$ of an object, we prompt the model to generate a description $D_B$ for that object by leveraging its ability of grounding caption. Due to the complexity of scenes or the presence of duplicate objects, the generated description may be noisy or fails to uniquely describe the corresponding object.
Then, the self-consistent filtering stage is adopted to filter those low-quality descriptions. Specifically, with the generated description $D_B$, we locate this description in the image and predict the bounding box $\hat{B}$ by leveraging the visual grounding ability of our model. The generated description will be removed if the Intersection Over Union (IOU) between $B$ and $\hat{B}$ is below a pre-defined threshold $\lambda$. This stage ensures that only high-quality descriptions are retained.

These two stages are performed to extend every annotated object in the dataset with textual description. This extended dataset is then well-suited for a wide range of RC tasks, \emph{e.g.}, coarse visual spatial reasoning, object detection, object counting, visual grounding and grounding caption. Illustrations of generated data are in Supplementary Material.
This self-consistent bootstrapping method serves as a powerful tool to harness the potential of object detection datasets for enhancing the RC ability.

\begin{figure}
    \begin{center}
    \includegraphics[width=1.0\linewidth]{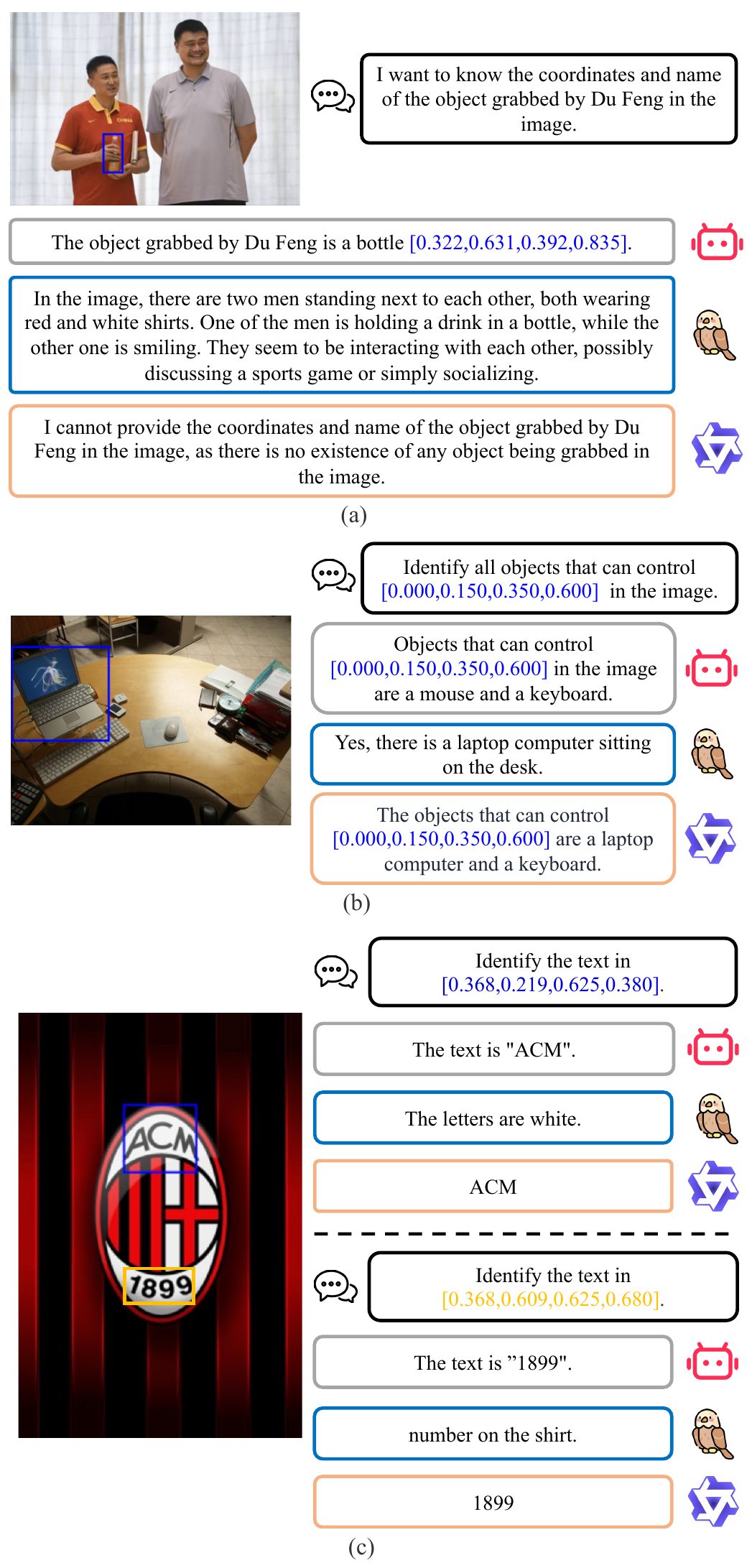}
    \end{center}
    \caption{A comparison of Pink \protect\logo~ against the recent MLLMs Shikra \protect\shikralogo~ and Qwen-VL \protect\qwen~on RC reasoning.}
    \label{fig:vis_main}
\end{figure}

\subsection{Qualitative Results on RC Reasoning}
In Fig.~\ref{fig:vis_main}, we compare some qualitative results on RC reasoning with Shikra~\cite{chen2023shikra} and Qwen-VL~\cite{bai2023qwen}. Pink exhibits substantially better capability on these RC tasks.

To answer the question as shown in Fig.~\ref{fig:vis_main} (a), the model needs to first identify Du Feng, a famous Chinese basketball player, then understand the action of grabbing. Pink successfully provides the correct answer. Pink also exhibit better reasoning capability as shown in Fig.~\ref{fig:vis_main} (b). It successfully identifies the referred region as laptop, and inferred that, the mouse and keyboard in the image could control the labtop. Shikra fails to follow the instruction and provides an un-related answer. Similarly, Qwen-VL only outputs the category of the referred object.

As shown in Fig.~\ref{fig:vis_main} (c), trained with millions of OCR-based data in the instruction tuning, Qwen-VL can give correct responses to OCR instruction. It is interesting to observe that, without any OCR-based data, our model also accurately recognizes characters located at referred region. It indicates that, Pink exhibits promising generalization capability to different RC tasks. More qualitative results can be found in Supplementary Material. Extensive experiments are conducted in following section.

\begin{table*}
    \small
     \setlength{\tabcolsep}{12.5px}
     \begin{center}
     \begin{tabular}{c|ccc|ccc}
     \hline
     Settings & IconQA & VSR & OK-VQA & RefCOCO\_val & Local & LookTwice \\ \hline
     Baseline & 44.6 & 65.6 & 58.5 & 55.0 & 0.0 & 0.2 \\
     w/o VG & 43.1 & 62.8 & 58.3 & - & - & - \\
     + R & 44.4 & 65.7 & 58.5 & 52.1 & 17.1 & 12.8 \\
     + R,S & 46.2 & 65.8 & 58.5 & 52.7 & 50.9 & 60.0 \\
     + R,S,C & 47.4 & 65.7 & 58.9 & 53.1 & 53.4 & 60.7 \\
     \rowcolor{gray!30} + R,S,C,D & \textbf{47.8} & 66.3 & 59.5 & 54.1 & 54.6 & 63.1 \\ \hline
     + R,S,C,D + Object365$\dagger$ & 44.6 & 65.9 & 58.7 & 73.8 & 52.1 & 69.2 \\
     \rowcolor{gray!30} + R,S,C,D + Object365 & 47.7 & \textbf{67.1} & \textbf{59.5} & \textbf{77.0} & \textbf{57.2} & \textbf{70.3} \\
     \hline \hline
     Freezing & 42.9 & 61.5 & 58.3 & 37.2 & 44.9 & 57.5 \\
     Full-tuning & 36.9 & 48.6 & 33.1 & 0.05 & 26.1 & 54.1 \\
     LoRA & 44.3 & 65.4 & 58.9 & \textbf{54.7} & \textbf{56.7} & 62.2 \\
     \rowcolor{gray!30} Our & \textbf{47.8} & \textbf{66.3} & \textbf{59.5} & 54.1 & 54.6 & \textbf{63.1} \\ \hline
     \end{tabular}
     \caption{Ablation study on instruction tuning dataset construction and training settings of visual encoder under a zero-shot setting. ``Baseline'' denotes leveraging Visual Genome by only performing visual grounding and grounding caption tasks. ``VG'' denotes Visual Genome. ``R'', ``S'', ``C'', and ``D''  denote the visual relation reasoning, coarse visual spatial reasoning, object counting and object detection tasks, respectively. $\dagger$ denotes generated referring-expression-bounding-box pairs in Object365 are not filtered with the self-consistent method. ``Freezing'' and ``Full-tuning'' denotes freezing the visual encoder and training the entire visual encoder, respectively. ``LoRA'' denotes using LoRA instead of the Adapter to perform parameter-efficient tuning.}
     \label{tab:dataset}
     \end{center}
 \end{table*}

\section{Experiments}
\subsection{Experimental Setting}
\noindent\textbf{Model architecture.} We employ the ViT-L/14~\cite{vit} as the visual encoder, which is pre-trained with CLIP~\cite{radford2021learning}. We choose an instruction-tuned model Vicuna-7B~\cite{vicuna} based on LLaMA-1~\cite{llama} as the LLM. The projection layer is a single fully connected layer. The Adapters are inserted before each self-attention layer of both the visual encoder and the LLM, with a hidden dimension $d_s=8$. The tunable parameter numbers of Adapter in the visual encoder and LLM are 393,216 and 2,097,152, respectively. The number of parameters in the projection layer is 4,194,304. Therefore, the total number of tunable parameters is about 6.7M.

\noindent\textbf{Training data.} The first stage utilizes 595K image-text pairs from CC3M~\cite{conceptualcap}, the same as LLaVA~\cite{llava}. The second stage adopts VQAv2~\cite{goyal2017making}, LLaVA-150K~\cite{llava}, A-OKVQA~\cite{schwenk2022okvqa}, Flickr30K~\cite{plummer2015flickr30k}, Visual Genome~\cite{krishna2017visual} and Object365~\cite{shao2019objects365} with referring-expression-bounding-box pairs generated by our self-consistent bootstrapping method. At each training iteration, when using an image in Visual Genome or Object365, one designed RC task will be selected randomly.
The model used to generate referring-expression-bounding-box pairs in Object365 is trained with the aforementioned datasets, excluding Object365 itself.
Note that we reduce the probability of sampling Object365 in batch construction to avoid a large number of training samples in Object365 dominating the training.

\noindent\textbf{Training details.} AdamW is adopted as the optimizer. In the first stage, the model is trained for 1 epoch with a batch size of 128 and weight decay of 0.0. After a warm-up period of 200 steps, the learning rate starts at 0.03 and decays to 0 with the cosine schedule. In the second stage, the model is trained for 6 epochs with a batch size of 32 and weight decay of 0.05. The warm-up phase consists of 10k steps and the learning rate starts at 5e-4. The input image is resized to $224 \times 224$ without any additional data-augmentation.
We set $\lambda$ as 0.5 to filter out low-quality descriptions.
The model is trained using 8 NVIDIA A100 GPUs. It takes about 1 and 30 hours for the first and second stage, respectively.

\noindent\textbf{Evaluation settings.} We evaluate our model on various datasets under the zero-shot and fine-tuning settings to validate the instruction-following ability of the trained model. These datasets encompass conventional multi-modal reasoning tasks, including conventional VQA (VQAv2~\cite{goyal2017making}, MMBench~\cite{liu2023mmbench}), abstract diagram understanding (IconQA~\cite{lu2021iconqa}), visual spatial reasoning (VSR~\cite{liu2023visual}), knowledge-intensive VQA (OK-VQA~\cite{marino2019ok}), scene understanding (GQA~\cite{hudson2019gqa}), and RC tasks such as RefCOCO/+~\cite{kazemzadeh2014referitgame}, RefCOCOg~\cite{mao2016generation}, Visual-7W~\cite{zhu2016visual7w}, PointQA-Local/LookTwice~\cite{mani2020point}.

\subsection{Ablation Study}
\noindent\textbf{Instruction tuning dataset construction.} To investigate the impact of instruction tuning dataset construction, we conduct ablation studies by excluding Visual Genome, designed RC tasks, and Object365 with referring-expression-bounding-box pairs. Results are summarized in Table~\ref{tab:dataset}. 

As shown in Table~\ref{tab:dataset}, enhancing the MLLM with RC task can benefit the conventional multi-modal reasoning tasks, \emph{e.g.}, removing visual grounding and grounding caption tasks leads to a performance degradation of 2.8\% on VSR.
When the model is trained solely with visual grounding and grounding caption, it fails to provide correct responses to the questions in PointQA-Local/LookTwice, indicating limited instruction-following ability for RC tasks.
As more RC tasks are included, the model begins to exhibit better instruction-following ability for these tasks. The performance on PointQA-Local/LookTwice increases from 0.0\%/0.2\% to 54.6\%/63.1\%.
The combination of all designed RC tasks yields the best performance on both conventional multi-modal reasoning tasks and RC tasks, thus validating the effectiveness of our instruction tuning dataset construction method.
We also observe that incorporating Object365 further enhances the performance of our method on RC tasks. For example, on RefCOCO\_val, the zero-shot accuracy increases from 54.1\% to 77.0\%. Notably, the exclusion of the self-consistent method results in the degradation of performance from 77.0\% to 73.8\% on RefCOCO\_val due to low-quality referring-expression-bounding-box pairs generated by the model. These results demonstrate the importance of the generated data and underscore the importance of the proposed self-consistent method.

\noindent\textbf{Training settings of visual encoder.} We further assess the effectiveness of different settings for training the visual encoder, and summarize results in Table~\ref{tab:dataset}.

Full-tuning the visual encoder results in a significant performance degradation. The performance on RefCOCO\_val drops from 54.1\% to 0.05\%.
This result aligns with the conclusion in~\cite{wang2023makes} that fine-tuning the visual encoder using a small-scale instruction tuning dataset can lead to a subsequent drop in performance.
Freezing the visual encoder also leads to performance degradation on various datasets. It can be attributed to the limited ability of the visual encoder to fine-grained image understanding. Our design allows for the optimization of both modalities and leverages the benefits of multi-modal instruction tuning, resulting in improved performance.
Moreover, using LoRA~\cite{lora} instead of the Adapter to perform parameter-efficient tuning can also achieve improved performance compared with Full-tuning or Freezing, demonstrating the effectiveness of adapting the visual encoder during multi-modal instruction tuning.

\subsection{Comparison with Recent Works}
This section proceeds to validate the effectiveness of our method through comparison with other recent works. 

\noindent\textbf{Evaluation on the conventional multi-modal reasoning tasks.}
To evaluate the instruction-following ability of our method, we conduct experiments of five multi-modal reasoning tasks on public benchmarks. These benchmarks and tasks assess various aspects of multi-modal comprehension ability of the model.
As shown in Table~\ref{tab:zero-shot-conventional}, our model consistently achieves the best performance using fewer trainable parameters, a smaller training set, and lower input image resolution. This demonstrates that the constructed instruction tuning dataset can effectively enhance the fine-grained perception capability.
%
\begin{table*}
    \small
    \begin{subtable}{1.0\linewidth}
        \setlength{\tabcolsep}{5.5px}
        \begin{center}
        \begin{tabular}{c|cccc|ccccc}
        \hline
        Models & Res. & \#PT Data & \#IT Data & \#Trainable Param. & VQAv2 & IconQA & VSR & OK-VQA & GQA \\ \hline
        Instruct-BLIP~\cite{dai2023instructblip} & 224 & 129M & 1.2M & 188M & - & 43.1 & 54.3 & - & 49.2 \\
        Shikra-7B~\cite{chen2023shikra} & 224 & 595K & 5.5M & 7B & 76.7$\dagger$ & 24.3 & 63.3 & 53.5 & 47.4 \\ \hline
        \rowcolor{gray!30} Pink & 224 & 595K & \textbf{396K} & \textbf{6.7M} & \textbf{78.7}$\dagger$ & \textbf{47.8} & \textbf{66.3} & \textbf{59.5} & \textbf{52.6} \\ \hline
        Qwen-VL~\cite{bai2023qwen} & 448 & 1.4B & 50M & 8B & 78.8$\dagger$ & - & - & 58.6$\dagger$ & 59.3$\dagger$ \\
        LLaVA-1.5~\cite{liu2023improved} & 336 & \textbf{558K} & 665K & 7B & 78.5$\dagger$ & - & - & - & 62.0$\dagger$ \\ \hline
        \rowcolor{gray!30} Pink+ & 224 & 595K & \textbf{477K} & \textbf{6.7M} & \textbf{78.8}$\dagger$ & 48.8 & 67.4 & \textbf{60.6}$\dagger$ & \textbf{64.5}$\dagger$ \\
        \hline
        \end{tabular}
        \captionsetup{font=small}
        \caption{Results on the conventional multi-modal reasoning tasks. $\dagger$ denotes the training set of corresponding dataset is included. + denotes adding the training set of OK-VQA and GQA during instruction tuning. ``Res.'', ``\#Trainable Param.'', ``\#PT Data'', and ``\#IT Data'' indicate input image resolution, the number of trainable parameters, the number of samples in pre-training and instruction tuning stage, respectively.}
        \label{tab:zero-shot-conventional}
        \end{center}
    \end{subtable}
 \vspace{0.3mm}

\begin{subtable}{1.0\linewidth}
    \setlength{\tabcolsep}{14.5px}
    \begin{center}
    \begin{tabular}{c|ccc|ccc|cc}
    \hline
    \multirow{2}{*}{Models} & \multicolumn{3}{c|}{RefCOCO} & \multicolumn{3}{c|}{RefCOCO+} & \multicolumn{2}{c}{RefCOCOg} \\ \cline{2-9}
    &    val   &  testA & testB & val & testA & testB & val & test \\ \hline
    Kosmos-2~\cite{peng2023kosmos} & 52.3 & 57.4 & 47.3 & 45.5 & 50.7 & 42.2 & 60.6 & 61.7 \\
    \rowcolor{gray!30} Pink & 54.1 & 61.2 & 44.2 & 43.9 & 50.7 & 35.0 & 59.1 & 60.1 \\
    \rowcolor{gray!30} Pink$^*$ & \textbf{77.0} & \textbf{82.4} & \textbf{68.2} & \textbf{65.6} & \textbf{75.2} & \textbf{53.4} & \textbf{72.4} & \textbf{74.0} \\ \hline
    \end{tabular}
    \captionsetup{font=small}
    \caption{Zero-shot results on the visual grounding task.}
    \label{tab:zero-shot-vg}
    \end{center}
\end{subtable}
 \vspace{0.3mm}

\begin{subtable}{1.0\linewidth}
    \setlength{\tabcolsep}{4.5px}
    \begin{center}
    \begin{tabular}{c|cc|ccc|ccc|cc|c|c}
    \hline
    \multirow{2}{*}{Models} & \multirow{2}{*}{Visual Encoder} & \multirow{2}{*}{Res.} & \multicolumn{3}{c|}{RefCOCO} & \multicolumn{3}{c|}{RefCOCO+} & \multicolumn{2}{c|}{RefCOCOg} & \multirow{2}{*}{Visual-7W} & \multirow{2}{*}{LookTwice}  \\ \cline{4-11}
    & & &   val   &  testA & testB & val & testA & testB & val & test &  & \\ \hline
    OFA-L~\cite{wang2022ofa} & ResNet152 & 480 & 80.0 & 83.7 & 76.4 & 68.3 & 76.0 & 61.8 & 67.6 & 67.6 & - & - \\
    Shikra-7B~\cite{chen2023shikra} & ViT-L & 224 & 87.0 & 90.6 & 80.2 & 81.6 & 87.4 & 72.1 & 82.3 & 82.2 & 84.3 & 72.1 \\ \hline
    \rowcolor{gray!30} Pink & ViT-L & 224 & 88.3 & 91.7 & 84.0 & 81.4 & 87.5 & 73.7 & 83.7 & 83.7 & 85.1 & 73.5 \\
    \rowcolor{gray!30} Pink$^{*}$ & ViT-L & 224 & \textbf{88.7} & \textbf{92.1} & \textbf{84.0} & \textbf{81.8} & \textbf{88.2} & \textbf{73.9} & \textbf{83.9} & \textbf{84.3} & \textbf{85.3} & \textbf{73.6} \\ \hline
    Qwen-VL~\cite{bai2023qwen} & ViT-G & 448 & 89.4 & 92.3 & 85.3 & 83.1 & 88.3 & 77.2 & 85.6 & 85.5 & - & - \\ \hline
    \rowcolor{gray!30} Pink-G & ViT-G & 224 & \textbf{91.5} & \textbf{93.4} & \textbf{88.0} & \textbf{86.0} & \textbf{89.5} & \textbf{79.8} & \textbf{86.8} & \textbf{87.8} & \textbf{86.8} & \textbf{76.6} \\
    \hline
    \end{tabular}
    \captionsetup{font=small}
    \caption{Fine-tuning results on the RC tasks. Pink-G indicates the ViT-G is used as the visual encoder for a fair comparison.}
    \label{tab:other_rc}
    \end{center}
\end{subtable}
 \vspace{0.3mm}
 
\begin{subtable}{1.0\linewidth}
    \setlength{\tabcolsep}{16px}
    \begin{center}
    \begin{tabular}{c|c|cccccc}
    \hline
    Models & Overall & LR & AR & RR & FP-S & FP-C & CP \\ \hline
    Kosmos-2~\cite{peng2023kosmos} & 58.2 & 48.6 & 59.9 & 34.7 & 65.6 & 47.9 & 70.4 \\
    LLaVA-1.5~\cite{liu2023improved} &  59.5 & 32.4 & 72.6 & 49.3 & 62.3 & 52.2 & 67.7 \\
    Qwen-VL~\cite{bai2023qwen} & 61.8 & 40.5 & 74.3 & 47.9 & 66.3 & 46.2 & 72.8 \\
    mPlug-Owl~\cite{ye2023mplug} & 68.5 & 56.8 & 77.9 & 62.0 & 72.0 & 58.4 & 72.6 \\ \hline
    \rowcolor{gray!30} Pink & \textbf{74.1} & \textbf{58.5} & \textbf{78.2} & \textbf{73.2} & \textbf{77.3} & \textbf{67.2} & \textbf{78.7} \\ \hline
    \end{tabular}
    \captionsetup{font=small}
    \caption{CircularEval results on MMBench test set~\citep{liu2023mmbench}.}
    \vspace{-3mm}
    \label{tab:mmbench}
    \end{center}
\end{subtable}
\caption{Comparison with other methods. * denotes Object365 with generated referring-expression-bounding-box pairs is used.}
\label{tab:zero-shot}
\vspace{-3mm}
\end{table*}

\noindent\textbf{Zero-shot evaluation on the visual grounding task.}
Visual grounding is a fundamental RC task that requires the model to predict the coordinates of a bounding box based on a given textual description.
We evaluate our model on three well-established datasets under the zero-shot setting in Table~\ref{tab:zero-shot-vg}. Our model significantly outperforms Kosmos-2~\cite{peng2023kosmos}, which is trained with a generated dataset GRIT containing 91M images and 115M referring expressions. This result validates the effectiveness of the proposed self-consistent bootstrapping method.

\noindent\textbf{Comparison with other models under fine-tuning setting on RC tasks.}
To further validate the RC ability of our method, we compare it with other models~\cite{wang2022ofa,chen2023shikra,bai2023qwen} that can perform RC tasks.
The comparison incorporates models that have the capability to handle various vision-language tasks. Models specifically designed for the visual grounding task are not included.
The instruction tuning dataset of compared models includes the training sets of datasets listed in Table~\ref{tab:other_rc}.
For a fair comparison, we also leverage those training sets in the instruction tuning of Pink. Note that, the training set size of Pink is still substantially smaller than those in compared methods, \emph{e.g.}, 50M of Qwen-VL~\cite{bai2023qwen} \emph{vs.} 519K of Pink.

The results in Table~\ref{tab:other_rc} show that our model obtains promising performance under the fine-tuning setting. This can be attributed to the diversity of RC tasks in the instruction tuning. Object365 further improves the performance, even training sets of these datasets are already included. This demonstrates the effectiveness of self-consistent bootstrapping method in converting existing dataset into more valuable RC training set. Qwen-VL utilizes a stronger visual encoder ViT-G trained by CLIP. To make a fair comparison, we change the visual encoder of Pink to ViT-G. Pink-G outperforms Qwen-VL by large margins, even with a lower input image resolution and less training samples. This result further validates the effectiveness of our training pipeline.

\noindent\textbf{Comparison with other models on MMBench test set.}
MMBench~\citep{liu2023mmbench} has been proposed as a new benchmark to evaluate various abilities of MLLMs, including logical reasoning (LR), attribute reasoning (AR), relation reasoning (RR), fine-grained perception single instance (FP-S), fine-grained perception cross instance (FP-C), and coarse perception (CP). We hence conduct experiments on MMBench to validate the capabilities of our model in all aspects.

The results summarized in Table~\ref{tab:mmbench} show that our method achieves the best overall performance among compared methods.
The main improvement comes from RR, FP-S, and FP-C. For example, compared with mPlug-Owl~\cite{ye2023mplug}, the accuracy boosting of Pink on LR and FP-C is 1.7\% and 8.8\%, respectively. These results demonstrate the strong fine-grained perception ability of Pink, which can be attributed to the incorporation of various RC tasks during the multi-modal instruction tuning.

\section{Conclusion}
\vspace{-1mm}
This paper presents a novel framework for enhancing fine-grained image perception ability of MLLMs. The framework includes a method for constructing an instruction tuning dataset by converting annotations from existing datasets into diverse RC tasks. A self-consistent bootstrapping method is proposed to extend object annotations to referring-expression-bounding-box pairs, enabling the acquisition of more instruction tuning data at a low cost. The visual encoder is tuned in a parameter-efficient way to gain fine-grained image understanding ability.
With fewer trainable parameters and less training data, our method achieves superior performance on both multi-modal tasks and RC tasks. 

\noindent\textbf{Acknowledgement} This work is supported in part by Natural Science Foundation of China under Grant No. U20B2052, 61936011, in part by the Okawa Foundation Research Award, and in part by Ant Group Research Intern Program.

\clearpage

{
    \small
    \bibliographystyle{ieeenat_fullname}
    \bibliography{main}
}

\clearpage
\setcounter{page}{1}
\setcounter{section}{1}
\renewcommand\thesection{\Alph{section}}
\maketitlesupplementary


\subsection{Discussion of Limitations}
Our method relies on the LLM. Therefore, it has some shortcomings from the LLM, such as bias or unfair response, and hallucination. Like most MLLMs with RC ability, we also find that our model is not good at object detection in complex scenarios, \emph{e.g.}, identifying multiple tiny objects in the image. It may be because the input resolution of the image is low. However, increasing the input resolution of the image is not a trivial task. Many efforts are still needed to deal with such tasks.

\subsection{Statistics of Training and Evaluation Datasets}\label{appendix:data}
The statistics for the training datasets and evaluation datasets are summarized in Table~\ref{tab:dataset_statistics} and Table~\ref{tab:evaluation_statistics}, respectively. We also provide an overview of the generated referring-expression-bounding-box pairs in Object365 in Table~\ref{tab:object365_statistics}. 
To ensure the quality of data, we apply certain filters during preprocessing. Firstly, we exclude images containing more than 15 objects. Moreover, for the purpose of bounding box description bootstrapping, we only consider objects that cover an area of more than 2,000 pixels. As a result, our dataset comprises 1,063,034 images, with a total of 4,961,822 generated referring-expression-bounding-box pairs. To further enhance the reliability of our dataset, we perform a self-consistent method that filters out 2,528,619 low-quality referring-expression-bounding-box pairs.

\begin{table}[]
    \small
    \begin{center}
    \begin{tabular}{c|cc}
    \hline
    Stage &  Dataset & Data Number \\ \hline
    PT & LLaVA-CC3M-Pretrain-595K~\citep{llava} & 595K \\ \hline
    IT & LLaVA-158K~\citep{llava} & 158K \\
    IT & VQAv2~\citep{goyal2017making} & 83K \\
    IT & A-OKVQA~\citep{schwenk2022okvqa} & 17K \\
    IT & Visual Genome~\citep{krishna2017visual} & 108K \\
    IT & Flickr30K~\citep{plummer2015flickr30k} & 30K \\ \hline
    IT & Object365$^*$~\citep{shao2019objects365} & 1M \\ \hline
    \end{tabular}
    \caption{Statistics of training datasets. PT and IT denote the pre-training and instruction tuning stage, respectively.}
    \label{tab:dataset_statistics}
    \end{center}
\end{table}

\begin{table}[]
    \small
    \setlength{\tabcolsep}{2.5px}
    \begin{center}
    \begin{tabular}{c|c|c}
    \hline
    Dataset & Split & Metric \\ \hline
    VQAv2~\cite{goyal2017making} & test-dev & VQA Score \\
    IconQA~\citep{lu2021iconqa} & multi-text-choice test & Accuracy \\
    VSR~\citep{liu2023visual} & zero-shot test & VQA Score \\
    OK-VQA~\citep{marino2019ok} & val & VQA Score \\
    GQA~\citep{hudson2019gqa} & test-dev & VQA Score \\  \hline
    RefCOCO~\citep{kazemzadeh2014referitgame} & val \& testA \& testB & Accuracy \\
    RefCOCO+~\citep{kazemzadeh2014referitgame} & val \& testA \& testB & Accuracy \\
    RefCOCOg~\citep{mao2016generation} & val \& test & Accuracy \\  \hline
    PointQA-Local~\citep{mani2020point} & test-dev & VQA Score \\
    PointQA-LookTwice~\citep{mani2020point} & test & VQA Score \\
    Visual-7W~\citep{zhu2016visual7w} & which box test & Accuracy \\ \hline
    MMBench~\citep{liu2023mmbench} & test & Accuracy \\ \hline
    \end{tabular}
    \caption{Summary of evaluation datasets.}
    \label{tab:evaluation_statistics}
    \end{center}
\end{table}

\begin{table}[]
    \small
    \begin{center}
    \begin{tabular}{ccc}
    \hline
    Images & Referring-expressions  & Avg Expression Length \\ \hline
    1,063,034 & 2,433,203 & 3.6 \\ \hline
    \end{tabular}
    \caption{Statistics of generated referring-expression-bounding-box pairs in Object365.}
    \label{tab:object365_statistics}
    \end{center}
\end{table}

\subsection{More experimental results}
The results that leverages a more advanced LLM LLaMA-2~\cite{touvron2023llama} are shown in Table~\ref{tab:llama2} and Table~\ref{tab:other_rc_llama2}. We can observe an improvement in performance on most datasets.

Similar to MMBench, SEED-Bench~\cite{li2023seed} is a recently proposed benchmark to evaluate the abilities of MLLMs on various evaluation dimensions including Scene Understanding (SU), Instance Identity (IId), Instance Attribute (IA), Instance Location (IL), Instance Counting (IC), Spatial Relation (SR), Instance Interaction (IIn), Visual Reasoning (VR), and Text Recognition (TR). We further conduct experiment on this benchmark. The results are shown in Table~\ref{tab:seed}. Pink also exhibits the best performance among compared MLLMs.
Our model has significant advantages in IIn, SR, and IL tasks, which require strong fine-grained image perception ability of MLLMs. These results further validate the effectiveness of our method to enhance the fine-grained perception ability of MLLMs.


\begin{table}
    \small
    \setlength{\tabcolsep}{2.5px}
    \begin{center}
    \begin{tabular}{c|ccccc}
    \hline
    Models & VQAv2 & IconQA & VSR & OK-VQA & GQA \\ \hline
    Pink & 78.7$\dagger$ & 47.8 & 66.3 & 59.5 & 52.6 \\ 
    Pink-LLaMA-2 & 78.8$\dagger$ & 49.1 & 67.9 & 60.2 & 52.0 \\ \hline
    \end{tabular}
    \captionsetup{font=small}
    \caption{Results on the conventional multi-modal reasoning tasks with LLaMA-2~\cite{touvron2023llama} as the LLM. $\dagger$ denotes the training set of corresponding dataset is included.}
    \label{tab:llama2}
    \end{center}
\end{table}

\begin{table*}
    \small
    \setlength{\tabcolsep}{4.5px}
    \begin{center}
    \begin{tabular}{c|ccc|ccc|cc|c|c}
    \hline
    \multirow{2}{*}{Models} & \multicolumn{3}{c|}{RefCOCO} & \multicolumn{3}{c|}{RefCOCO+} & \multicolumn{2}{c|}{RefCOCOg} & \multirow{2}{*}{Visual-7W} & \multirow{2}{*}{LookTwice}  \\ \cline{2-9}
    &  val   &  testA & testB & val & testA & testB & val & test &  & \\ \hline
    Pink & 88.3 & 91.7 & 84.0 & 81.4 & 87.5 & 73.7 & 83.7 & 83.7 & 85.1 & 73.5 \\
    Pink-LLaMA-2 & 89.0 & 92.1 & 84.6 & 82.6 & 88.3 & 74.5 & 83.8 & 84.4 & 85.1 & 73.8 \\
    \hline
    \end{tabular}
    \captionsetup{font=small}
    \caption{Fine-tuning results on RC tasks with LLaMA-2~\cite{touvron2023llama} as the LLM.}
    \label{tab:other_rc_llama2}
    \end{center}
\end{table*}

\begin{table*}[]
    \small
    \begin{center}
    \begin{tabular}{c|c|ccccccccc}
    \hline
    Models & Overall & SU & IId & IA & IL & IC & SR & IIn & VR & TR \\ \hline 
    Kosmos-2~\cite{peng2023kosmos} & 54.4 & 63.4 & 57.1 & 58.5 & 44.0 & 41.4 & 37.9 & 55.7 & 60.7 & 25.9 \\
    Instruct-BLIP~\cite{dai2023instructblip} & 58.8 & 60.2 & 58.9 & 65.6 & 43.6 & \textbf{57.2} & 40.3 & 52.6 & 47.7 & 43.5 \\
    mPlug-Owl~\cite{ye2023mplug} & 37.9 & 49.7 & 45.3 & 32.5 & 36.7 & 27.3 & 32.7 & 44.3 & 54.7 & 28.8 \\
    Qwen-VL~\cite{bai2023qwen} & 62.3 & 71.2 & 66.4 & 67.7 & 53.5 & 44.8 & 43.8 & 62.9 & \textbf{74.9} & 51.2 \\ \hline
    \rowcolor{gray!30} Pink & 66.2 & 73.1 & 69.1 & 69.1 & 60.5 & 55.2 & \textbf{51.0} & \textbf{76.3} & 70.4 & \textbf{58.8} \\
    \rowcolor{gray!30} Pink-LLaMA-2 & \textbf{67.0} & \textbf{75.2} & \textbf{70.1} & \textbf{70.1} & \textbf{63.3} & 53.8 & 50.2 & 69.1 & 74.3 & 50.0 \\ \hline
    \end{tabular}
    \caption{Results on the image split of SEED-Bench~\cite{li2023seed}.}
    \label{tab:seed}
    \end{center}
\end{table*}

\subsection{Qualitative Analysis}\label{appendix:vis}
We show qualitative results on different types of vision-language tasks and RC tasks. The results of mPlug-Owl~\cite{ye2023mplug}, Shikra~\cite{chen2023shikra} and Qwen-VL~\cite{bai2023qwen} are from their official online demos.

\begin{figure}
    \begin{center}
    \includegraphics[width=1.0\linewidth]{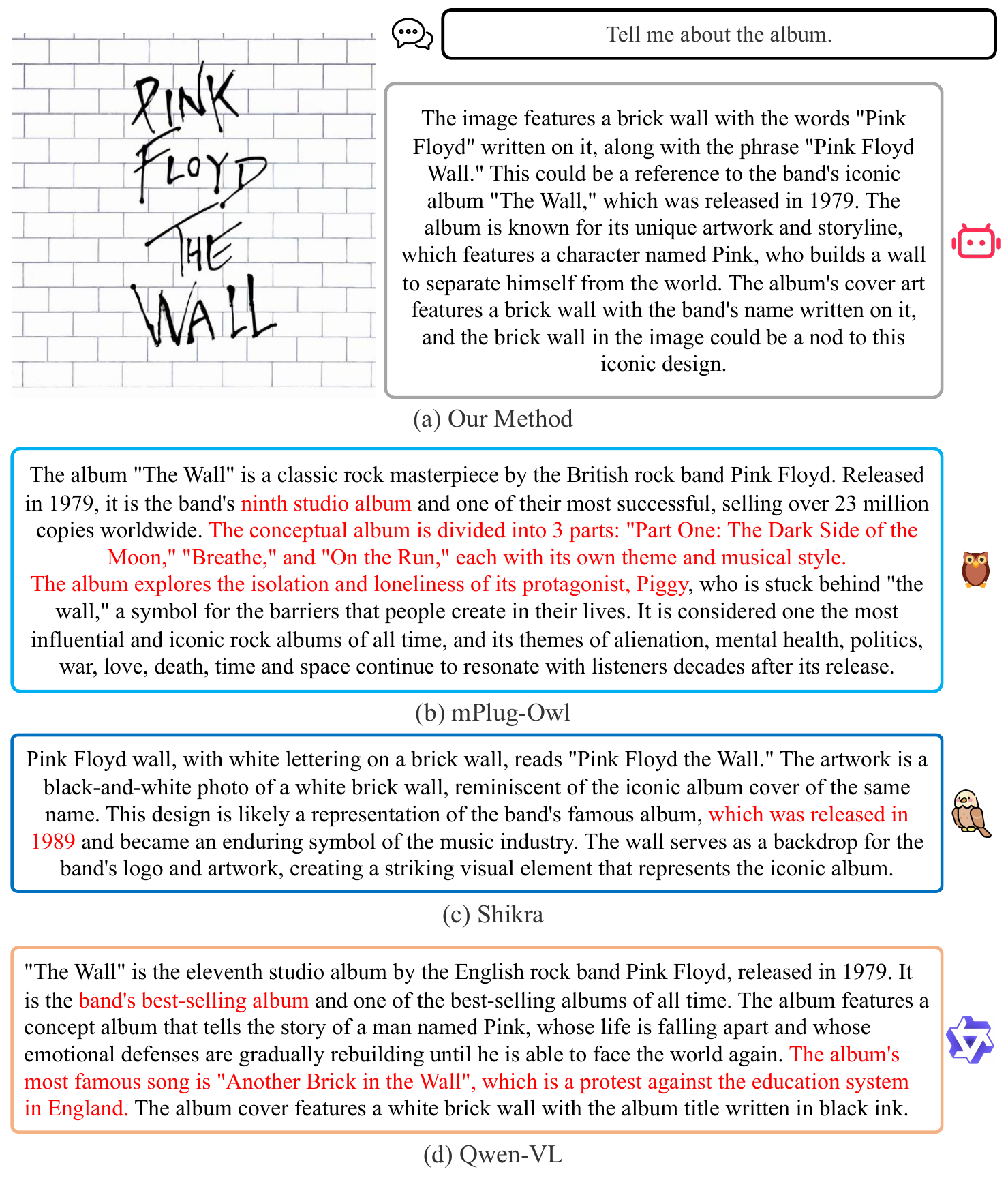}
    \end{center}
    \caption{A comparison of knowledge-intensive QA. We mark the hallucination part in \textcolor{red}{red}.}
    \label{fig:vis_image1}
\end{figure}

\noindent\textbf{Knowledge-intensive QA.} Fig~\ref{fig:vis_image1} depicts an example of this task, where the model is required to identify an album cover and provide an introduction for the album. mPlug-Owl and Qwen-VL can identify the album $\emph{The Wall}$ correctly but begins to make hallucination when introducing this album. Some introductions provide by mPlug-Owl do not align with the actual facts associated with the album.
The answer of Shikra focuses on the content of the image, indicating a poor instruction-following ability of Shikra.
In contrast, our model not only correctly identifies the album but also delivers an accurate and relevant introduction, showcasing its strong instruction-following capability.

\begin{figure}
    \begin{center}
    \includegraphics[width=1.0\linewidth]{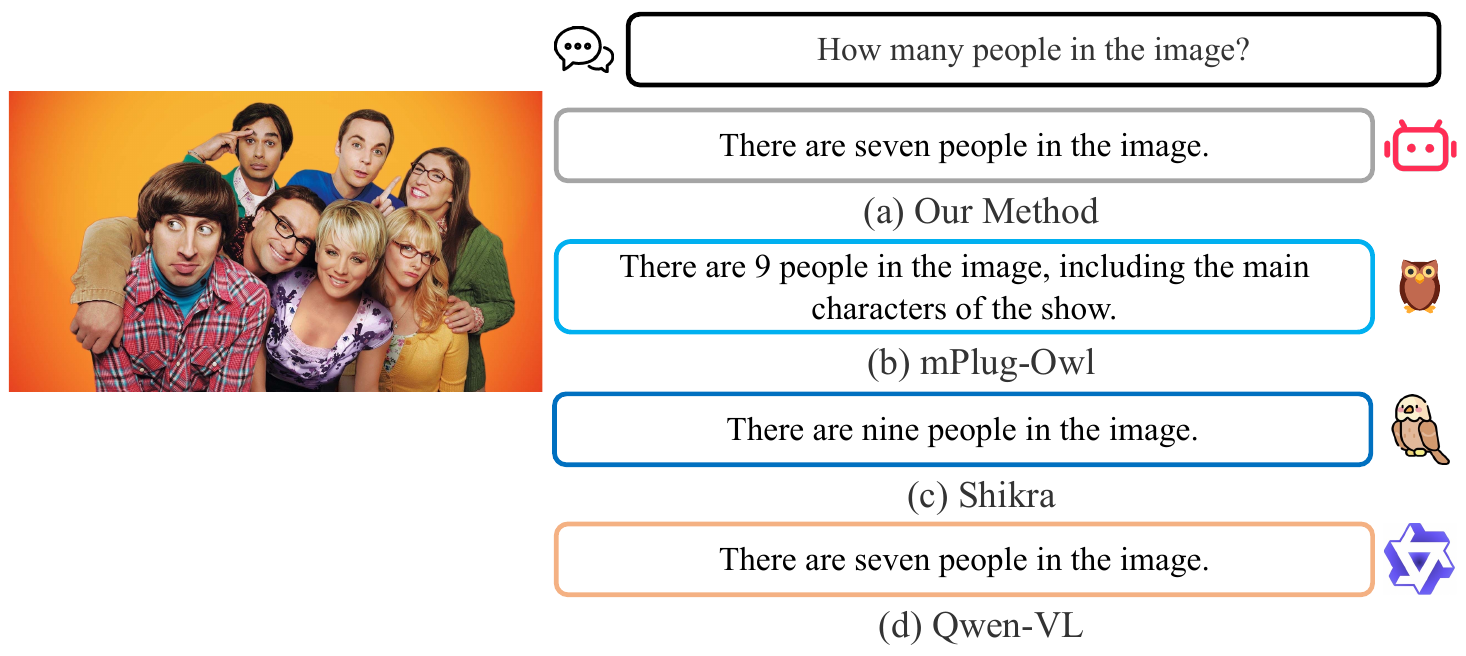}
    \end{center}
    \caption{A comparison of fine-grained QA.}
    \label{fig:vis_image2}
\end{figure}

\noindent\textbf{Fine-grained QA.} As shown in Fig.~\ref{fig:vis_image2}, mPlug-Owl fails to output the correct number of people in the image. It may be because that its lack of RC ability hinders its fine-grained understanding of images.
Despite Shikra has RC ability, it also fails to give a correct answer. This result can validate that trained with limited RC tasks Shikra struggles to achieve a fine-grained perception of images.
In contrast, our model makes a correct answer to the question, demonstrating its strong capability in fine-grained image perception.

\begin{figure}
    \begin{center}
    \includegraphics[width=1.0\linewidth]{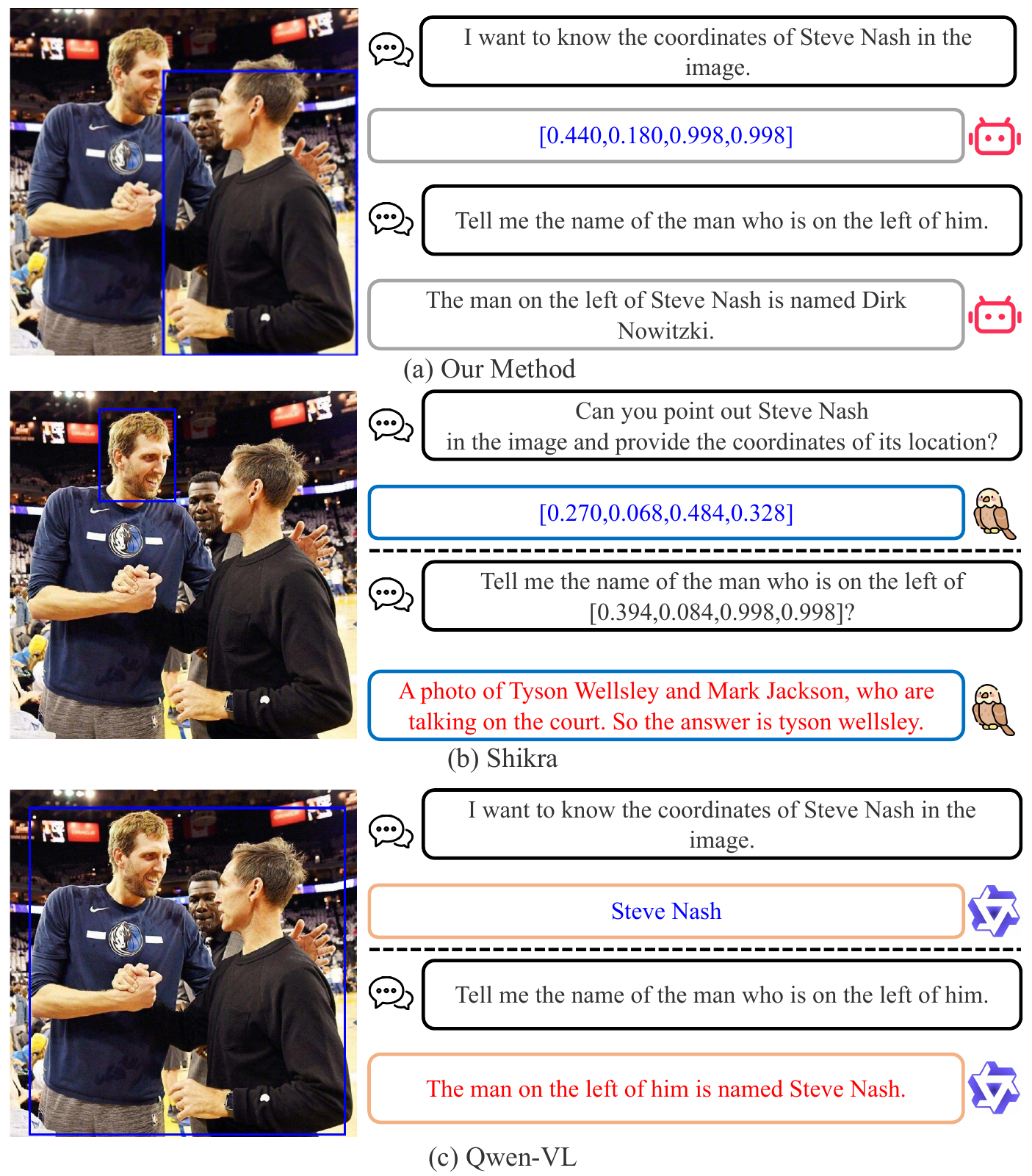}
    \end{center}
    \caption{A comparison of referential comprehension reasoning. We mark the hallucination part in \textcolor{red}{red}.}
    \label{fig:vis_rc}
\end{figure}

\noindent\textbf{Referential comprehension reasoning.} We show a case of RC reasoning in Fig.~\ref{fig:vis_rc}. In this case, our model accurately locates Steve Nash. Shikra seems to be unaware of who is Steve Nash. Surprisingly, Shikra's output coordinates point to the face of Dirk Nowitzki. We also need to point out the result of Shikra is generated using the instruction template that is identical to the one used during training. Shikra shows limited instruction-following ability when using a template that is different from the one used during training.
Our model also shows a ability for multi-round conversation. It can understand what ``him'' refers to in the second question. As Shikra makes a wrong response for the first question, in the second question, we give the correct coordinates manually. However, Shikra fails to follow the instructions and begins to hallucination.
Qwen-VL fails to locate Steve Nash correctly. In the multi-round conversation, it also gives a wrong response.
These results can further validate the RC ability of our model.

\begin{figure}
    \begin{center}
    \includegraphics[width=1.0\linewidth]{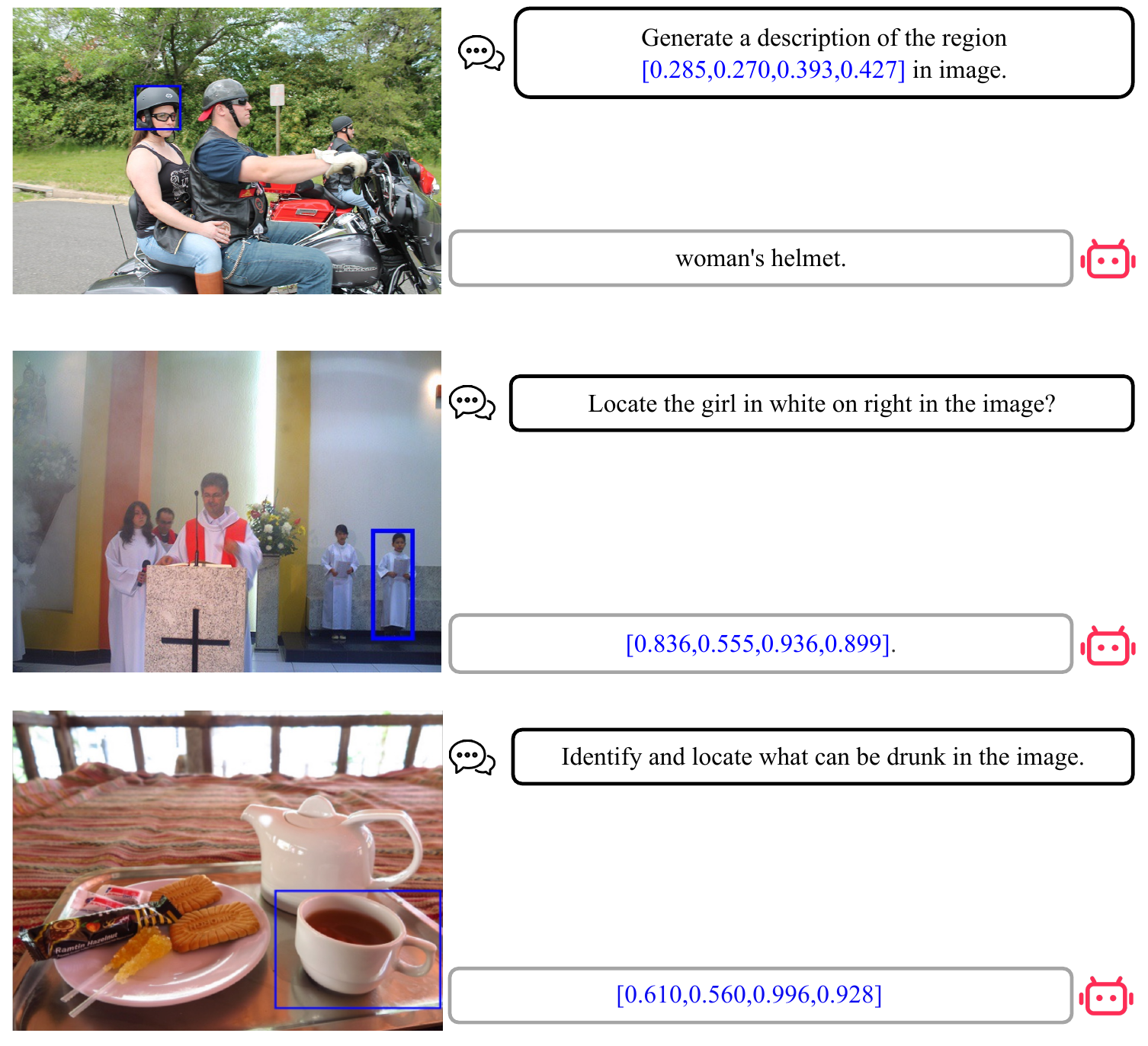}
    \includegraphics[width=1.0\linewidth]{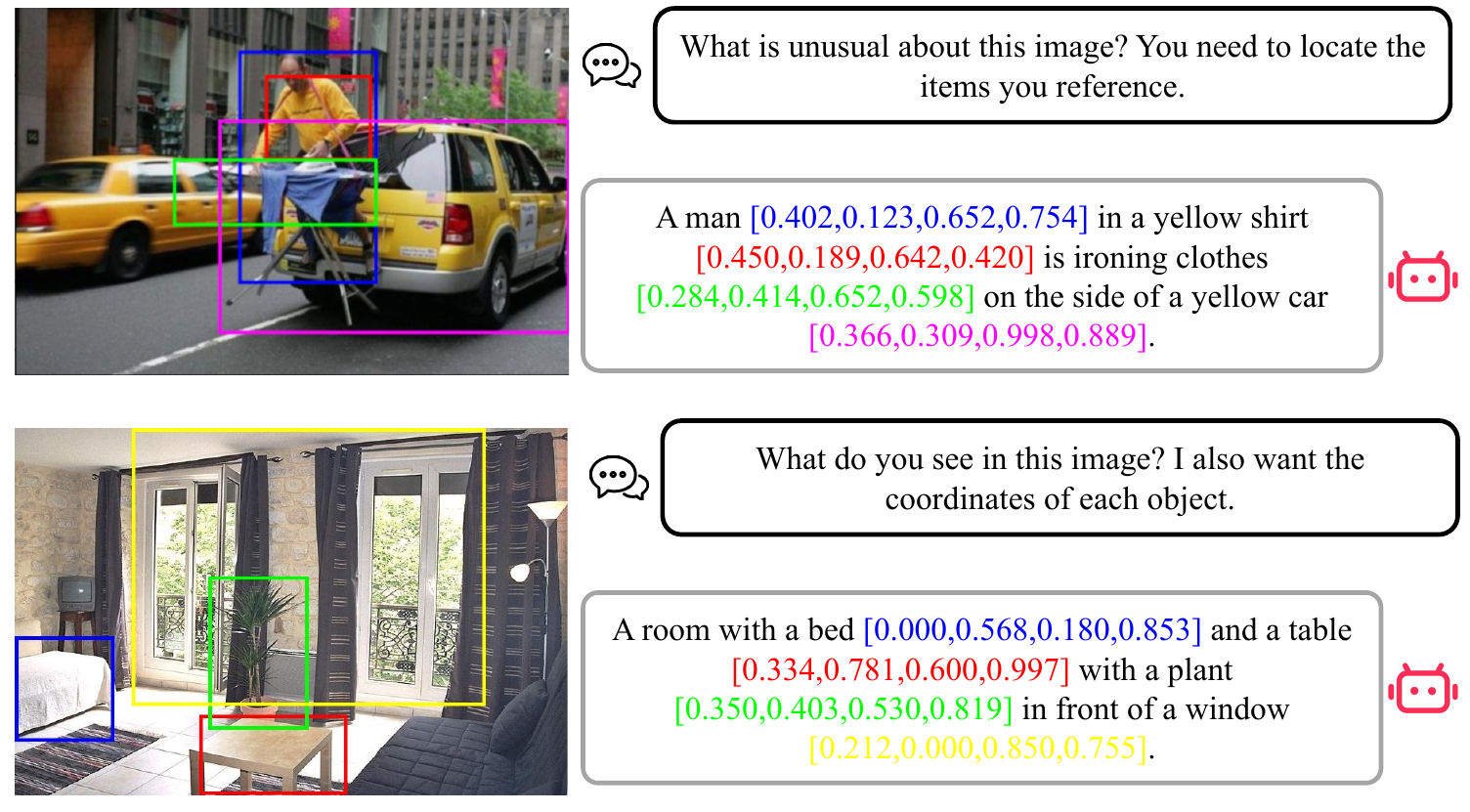}
    \end{center}
    \caption{More qualitative results of our model.}
    \label{fig:vis_more}
\end{figure}

\begin{figure}
    \begin{center}
    \includegraphics[width=1.0\linewidth]{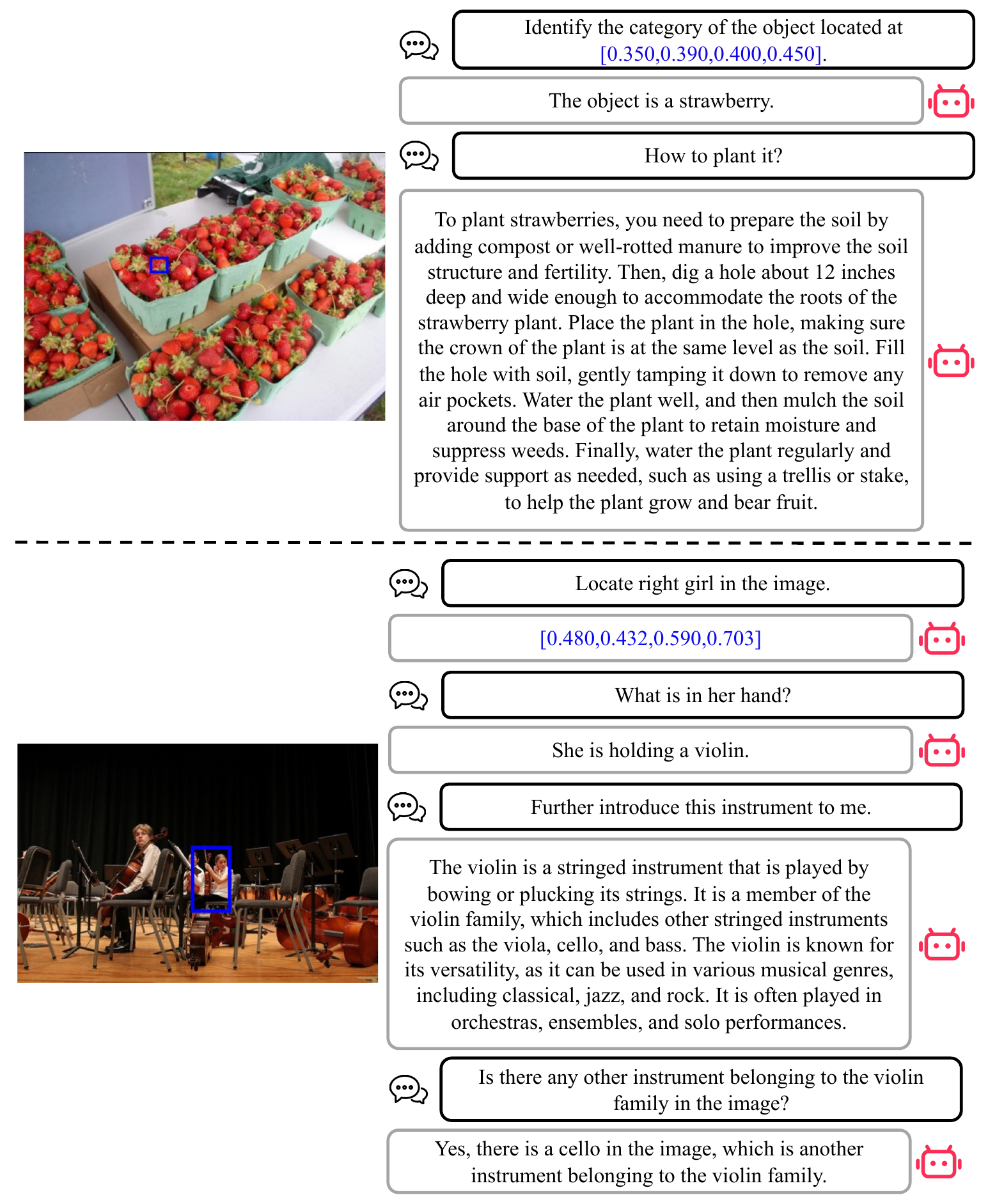}
    \end{center}
    \caption{Qualitative results of our model on multi-round conversation.}
    \label{fig:multi-turn}
\end{figure}

\begin{figure}
    \begin{center}
    \includegraphics[width=1.0\linewidth]{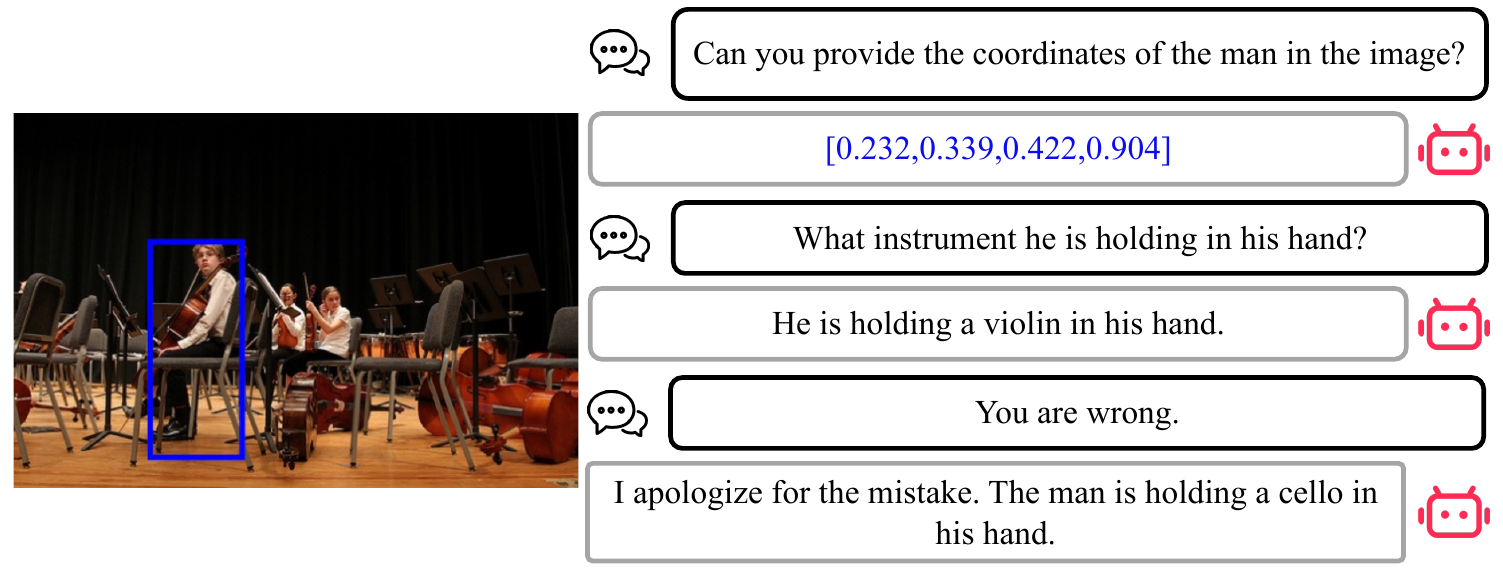}
    \end{center}
    \caption{Qualitative results of our model on multi-round conversation with correction.}
    \label{fig:multi-turn-correct}
\end{figure}

\begin{figure}
    \begin{center}
    \includegraphics[width=1.0\linewidth]{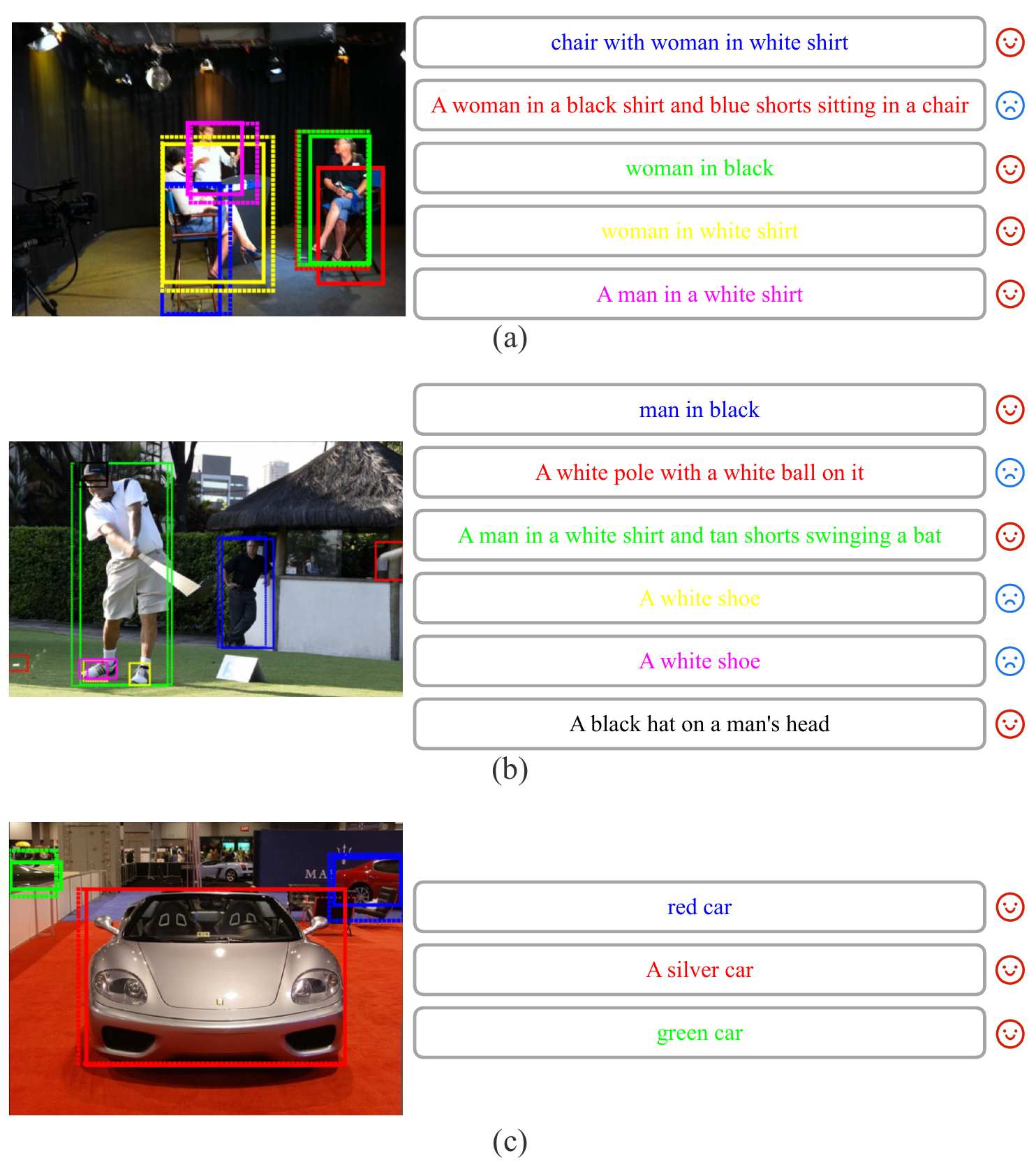}
    \end{center}
    \caption{Qualitative results of generated referring-expression-bounding-box pairs. The solid rectangle represents the ground-truth bounding box. The dashed rectangle represents the bounding box generated by our model with visual grounding according to the generated description. \protect\sad~denotes the generated description is filtered out by the proposed self-consistent method.}
    \label{fig:vis_object365}
\end{figure}

\noindent\textbf{More qualitative results.} More qualitative results of our model are shown in Fig.~\ref{fig:vis_more}. Our model demonstrates the ability to generate unique descriptions with contextual information when provided with coordinates of a specific area. For instance, instead of simply outputting ``helmet'', our model uses ``woman's'' features for differentiation.
Moreover, our model successfully locates items mentioned in descriptions that require outside knowledge. For example, it can correctly identify what can be drunk in the image.
In the last two cases, our model not only provides correct answers but also locates the mentioned items in the image. This ability can achieve more applications.

We also present qualitative results of our model on multi-round conversation about RC in Fig.~\ref{fig:multi-turn} and Fig.~\ref{fig:multi-turn-correct}. Our method can understand the complex referential relationships in the dialogue context, \emph{e.g.}, it, her and this instrument.
As shown in Fig.~\ref{fig:multi-turn-correct}, although the model initially gives the wrong answer about what is holding by the man, when it is told that its answer is incorrect, the model can make corrections and provide the correct answer. This result can further demonstrate the promising instruction-following ability of our model.
Additionally, these qualitative results highlight that the integration of RC ability significantly expands the range of tasks our model can successfully handle, thereby broadening its application potential.

\noindent\textbf{Qualitative results of generated referring-expression-bounding-box pairs.}\label{appendix:object365}
We illustrate some examples of generated referring-expression-bounding-box pairs in Fig.~\ref{fig:vis_object365}. For each object in the image, the bounding box description bootstrapping method can generate a description related to that object. Most of generated descriptions are correct. However, some generated descriptions exhibit incorrect or ambiguous descriptions that fail to uniquely identify an object. As shown in Fig.~\ref{fig:vis_object365} (a) and (b), when leveraging visual grounding to locate these descriptions in the image, IOU between the predicted bounding box and the ground-truth bounding box is low. Our self-consistent filtering method can effectively filter them out. These results can further validate the effectiveness of our self-consistent bootstrapping method to extend the object annotations to high-quality referring-expression-bounding-box pairs.

\subsection{Instruction Templates}\label{appendix:instruct}
\noindent\textbf{Instruction templates of visual relation reasoning.}
We list all the instruction templates of visual relation reasoning task below. For the Task2, we use different instruction templates to ask the model output different contents (coordinates, class name, or both coordinates and class name).

\begin{tcolorbox}[title=\small{Instruction templates of visual relation reasoning-Task1}]
    \small
    What is the relation between \textless subject\textgreater~and \textless object\textgreater? \\
    Describe the relation between \textless subject\textgreater~and \textless object\textgreater? \\
    Assist me in finding the relation between \textless subject\textgreater~and \textless object\textgreater~in the photo. \\
    In the given image, could you find and tell me the relation between \textless subject\textgreater~and \textless object\textgreater? \\
    I need help identifying the relation between \textless subject\textgreater~and \textless object\textgreater. Can you point it out in this image? \\
    What is the relation between \textless subject\textgreater~and \textless object\textgreater~in this picture? \\
    Could you describe the relation between \textless subject\textgreater~and \textless object\textgreater~in this image? \\
    I'm having trouble identifying the relation between \textless subject\textgreater~and \textless object\textgreater. Could you clarify it for me in this image? \\
    Can you help me understand the relationship between \textless subject\textgreater~and \textless object\textgreater~in this image? \\
    I'm trying to understand the relation between \textless subject\textgreater~and \textless object\textgreater. Can you help me by describing it? \\
    I need some assistance in identifying the relation between \textless subject\textgreater~and \textless object\textgreater~in this image. \\
    In this image, can you find and describe the relation between \textless subject\textgreater~and \textless object\textgreater~for me? \\
    Could you please explain the relation between \textless subject\textgreater~and \textless object\textgreater.
\end{tcolorbox}

\begin{tcolorbox}[title=\small{Instruction templates of visual relation reasoning-Task2\#1}]
    \small
    Assist me in locating the position of all the objects \textless relation\textgreater~the \textless subject\textgreater? \\
    I want to know the coordinates of all the objects \textless relation\textgreater~the \textless subject\textgreater? \\
    Detect all the objects have a relationship \textless relation\textgreater~with the \textless subject\textgreater~and output there locations. \\
    There are some objects that are \textless relation\textgreater~the \textless subject\textgreater. Could you tell me there locations? \\
    Identify all the objects that have a relationship \textless relation\textgreater~with the \textless subject\textgreater. Where are they located? \\
    Please locate all the objects that are \textless relation\textgreater~the \textless subject\textgreater~and provide their coordinates. \\
    Find all the objects that have a relation of \textless relation\textgreater~with the \textless subject\textgreater. Can you give me their positions? \\
    Point out the objects that are \textless relation\textgreater~the \textless subject\textgreater. Where can I find them? \\
    I need to locate all the objects that are \textless relation\textgreater~the \textless subject\textgreater. Can you assist me with this task? \\
    Could you help me find all the objects that have a relation of \textless relation\textgreater~with the \textless subject\textgreater? Please provide their locations. \\
    Please detect all the objects that are \textless relation\textgreater~the \textless subject\textgreater. Output their positions. \\
    Identify and provide the coordinates of all objects that are \textless relation\textgreater~the \textless subject\textgreater. \\
    Find the objects that have a relation of \textless relation\textgreater~with the \textless subject\textgreater. Where are they situated? \\
    What objects have the relation of \textless relation\textgreater~with the \textless subject\textgreater? Could you locate them for me? \\
    Can you help me locate all the objects that are \textless relation\textgreater~the \textless subject\textgreater~and give me their positions? \\
    Output the positions of all objects that have a relation of \textless relation\textgreater~with the \textless subject\textgreater. \\
    Identify the objects that are \textless relation\textgreater~the \textless subject\textgreater. Where are they located? \\
    Please locate all the objects that are \textless relation\textgreater~the \textless subject\textgreater~and provide their positions.
\end{tcolorbox}

\begin{tcolorbox}[title=\small{Instruction templates of visual relation reasoning-Task2\#2}]
    \small
    Assist me in identifying the categories of all the objects \textless relation\textgreater~by the \textless subject\textgreater? \\
    Detect all the objects \textless relation\textgreater~by the \textless subject\textgreater~and output there categories, respectively. \\
    There are some objects that are \textless relation\textgreater~the \textless subject\textgreater. Could you tell me there categories? \\
    I want to know the categories of all the objects whose relation is \textless relation\textgreater~with the \textless subject\textgreater? \\
    Identify the object categories that are \textless relation\textgreater~the \textless subject\textgreater. \\
    Find all objects that are related to \textless subject\textgreater~using the relationship \textless relation\textgreater, and categorize them. \\
    Your task is to recognize and classify all objects that are \textless relation\textgreater~by the \textless subject\textgreater. \\
    Please determine the categories of all objects that are \textless relation\textgreater~by the \textless subject\textgreater. \\
    Can you identify the categories of objects \textless relation\textgreater~the \textless subject\textgreater? \\
    Your job is to identify all objects that \textless relation\textgreater~the \textless subject\textgreater~and list their categories. \\
    Detect and categorize all objects that are \textless relation\textgreater~the \textless subject\textgreater. \\
    I need you to determine the categories of all objects that \textless relation\textgreater~the \textless subject\textgreater. \\
    Identify and classify all objects that are \textless relation\textgreater~the \textless subject\textgreater. \\
    Please identify the categories of all objects that are \textless relation\textgreater~the \textless subject\textgreater. \\
    Please help me identify the object categories whose relationship is \textless relation\textgreater~with \textless subject\textgreater. \\
    Recognize and categorize all objects that are related to \textless subject\textgreater~using the relationship \textless relation\textgreater. \\
    I need you to categorize all objects that is related to \textless subject\textgreater~with relationship as \textless relation\textgreater.
\end{tcolorbox}

\begin{tcolorbox}[title=\small{Instruction templates of visual relation reasoning-Task2\#3}]
    \small
    Your task is to locate all objects that have a relation \textless relation\textgreater~with \textless subject\textgreater~and classify them. \\
    I need you to categorize and locate all objects that is related to \textless subject\textgreater~with relationship as \textless relation\textgreater. \\
    Please locate and categorize all the objects that have a relation of \textless relation\textgreater~with \textless subject\textgreater. \\
    Assist me in locating and classifying all the objects \textless relation\textgreater~the \textless subject\textgreater? \\
    Find all the objects that have a relation of \textless relation\textgreater~with the \textless subject\textgreater. Can you give me their positions and categories \\
    Your task is to locate all objects that have a relation \textless relation\textgreater~with \textless subject\textgreater~and classify them. \\
    I need you to categorize and locate all objects that is related to \textless subject\textgreater~with relationship as \textless relation\textgreater. \\
    Please locate and categorize all the objects that have a relation of \textless relation\textgreater~with \textless subject\textgreater. \\
    Assist me in locating and classifying the position of all the objects \textless relation\textgreater~the \textless subject\textgreater? \\
    Find all the objects that have a relation of \textless relation\textgreater~with the \textless subject\textgreater. Can you give me their positions and categories? \\
    Your task is to locate and classify all objects that are related to \textless subject\textgreater~using the relationship \textless relation\textgreater. \\
    I need you to locate and categorize all objects having a relationship \textless relation\textgreater~with the given \textless subject\textgreater. \\
    Find all objects related to \textless subject\textgreater~with the relationship \textless relation\textgreater. Categorize and locate them for me. \\
    Your objective is to locate and classify the objects that are related to \textless subject\textgreater~through the relationship \textless relation\textgreater. \\
    I require you to detect and categorize all objects that have a relationship \textless relation\textgreater~with \textless subject\textgreater. \\
    Please find and classify all objects that has a relationship \textless relation\textgreater~with \textless subject\textgreater. \\
    Assist me in locating and categorizing all objects that related to \textless subject\textgreater~with the relationship \textless relation\textgreater. \\
    Find all objects that are related to \textless subject\textgreater~using the relationship \textless relation\textgreater. Categorize and locate their positions. \\
    Your task is to identify and classify all objects related to \textless subject\textgreater~through the relationship \textless relation\textgreater. \\
    I need you to locate and categorize all objects that have a relationship \textless relation\textgreater~with \textless subject\textgreater. \\
    Assist me in locating and classifying all objects that are related to \textless subject\textgreater~through the relationship \textless relation\textgreater. \\
    Find all the objects that has a relationship \textless relation\textgreater~with \textless subject\textgreater. Categorize and locate their positions for me.
\end{tcolorbox}

\noindent\textbf{Instruction templates of coarse visual spatial reasoning.}
We list all the instruction templates of coarse visual spatial reasoning task below. Similar to visual relation reasoning, different instruction templates are used to ask the model output different contents.

\begin{tcolorbox}[title=\small{Instruction templates of coarse visual spatial reasoning\#1}]
    \small
    Identify the objects located at \textless loc\textgreater~of \textless object\textgreater. Please classify them by category and provide their locations. \\
    I need to know what objects are present at \textless loc\textgreater~of \textless object\textgreater. Can you help me locate and categorize them? \\
    Find all the objects at \textless loc\textgreater~of \textless object\textgreater. Please provide me with their categories and locations. \\
    I want to know the categories and positions of the objects located at \textless loc\textgreater~of \textless object\textgreater.\\
    Locate and classify all the objects at \textless loc\textgreater~of \textless object\textgreater. \\
    Could you tell me the categories and positions of the objects present at \textless loc\textgreater~of \textless object\textgreater? \\
    Help me locate and categorize all the objects at \textless loc\textgreater~of \textless object\textgreater. \\
    I need to know the categories and locations of the objects at \textless loc\textgreater~of \textless object\textgreater. \\
    What are the categories and positions of the objects located at \textless loc\textgreater~of \textless object\textgreater? \\
    Identify and locate all the objects at \textless loc\textgreater~of \textless object\textgreater. I need their categories and positions. \\
    I want to know the categories and positions of the objects at \textless loc\textgreater~of \textless object\textgreater. \\
    Locate and classify all the objects at \textless loc\textgreater~of \textless object\textgreater. Please provide me with their categories and positions.
\end{tcolorbox}

\begin{tcolorbox}[title=\small{Instruction templates of coarse visual spatial reasoning\#2}]
    \small
    What are the categories of the objects located at \textless loc\textgreater~of \textless object\textgreater? \\
    Detect and classify all the objects at \textless loc\textgreater~of \textless object\textgreater. I need to know their categories. \\
    Please find and categorize all the objects present at \textless loc\textgreater~of \textless object\textgreater. \\
    Give the categories of all the objects you can find at \textless loc\textgreater~of \textless object\textgreater. \\
    I need you to find and categorize all the objects that are at \textless loc\textgreater~of \textless object\textgreater. \\
    Please provide me with the categories of all the objects present at \textless loc\textgreater~of \textless object\textgreater. \\
    What types of objects are located at \textless loc\textgreater~of \textless object\textgreater? Please list their categories. \\
    Please find all the objects at \textless loc\textgreater~of \textless object\textgreater~and give me their categories. \\
    What are the categories of the objects that are present at \textless loc\textgreater~of \textless object\textgreater? \\
    I need you to classify all the objects located at \textless loc\textgreater~of \textless object\textgreater. \\
    Please give me the categories of all the objects that are located at \textless loc\textgreater~of \textless object\textgreater.
\end{tcolorbox}

\begin{tcolorbox}[title=\small{Instruction templates of coarse visual spatial reasoning\#3}]
    \small
    What are the coordinates of the objects located at \textless loc\textgreater~of \textless object\textgreater? \\
    Detect and give the coordinates of all the objects at \textless loc\textgreater~of \textless object\textgreater. \\
    Please find and locate all the objects present at \textless loc\textgreater~of \textless object\textgreater. \\
    Give the detail locations of all the objects you can find at of \textless loc\textgreater~\textless object\textgreater. \\
    Locate all the objects and give there coordinates found at \textless loc\textgreater~of \textless object\textgreater. \\
    What are the positions of all the objects at \textless loc\textgreater~of \textless object\textgreater? \\
    Can you find and list the positions of all the objects present at \textless loc\textgreater~of \textless object\textgreater? \\
    Provide the coordinates of objects located at \textless loc\textgreater~of \textless object\textgreater. \\
    List and indicate the positions of all objects at \textless loc\textgreater~of \textless object\textgreater. \\
    Enumerate and specify the positions of all objects found at \textless loc\textgreater~of \textless object\textgreater. \\
    What objects are situated at \textless loc\textgreater~of \textless object\textgreater~and where precisely are they located? \\
    What are the coordinates of all objects found at \textless loc\textgreater~of \textless object\textgreater?
\end{tcolorbox}

\noindent\textbf{Instruction templates of object counting.}
We list all the instruction templates of object counting task below. \textless category\textgreater~will be replaced by the category name.

\begin{tcolorbox}[title=\small{Instruction templates of object counting\#1}]
    \small
    Can you tell me how many \textless category\textgreater~are present in this picture? \\
    I need to know the number of \textless category\textgreater~in this image. \\
    Count how many \textless category\textgreater~are in this picture. \\
    Please determine the quantity of \textless category\textgreater~shown in this image. \\
    How many instances of \textless category\textgreater~can you find in this picture? \\
    I would like to know how many \textless category\textgreater~are visible in this image. \\
    Count the number of \textless category\textgreater~that you see in this picture. \\
    Please provide me with the count of \textless category\textgreater~in this image. \\
    How many objects of \textless category\textgreater~are in this image? \\
    Can you count the items of \textless category\textgreater~in this picture? \\
    What is the total number of \textless category\textgreater~in this image? \\
    How many \textless category\textgreater~can you spot in this image? \\
    Please determine the quantity of \textless category\textgreater~in this image. \\
    Count the number of \textless category\textgreater~that appear in this picture. \\
    How many \textless category\textgreater~are in the picture? \\
    Counting the number of \textless category\textgreater~appeared in the image. \\
    Please give me the number of \textless category\textgreater~appeared in the image.
\end{tcolorbox}

\begin{tcolorbox}[title=\small{Instruction templates of object counting\#2}]
    \small
    How many objects in the image are of the same category as \textless object\textgreater? \\
    Count the number of objects in the image that are similar to \textless object\textgreater~in category. \\
    What is the total count of objects that share the same category as \textless object\textgreater~in the image? \\
    How many objects in the image have the same category as \textless object\textgreater? \\
    Count all the objects in the image that fall under the same category as \textless object\textgreater. \\
    What is the number of objects that share the same category as \textless object\textgreater~in the image? \\
    Count the objects that belong to the same category as \textless object\textgreater~in the image. \\
    How many objects of the same category as the object represented by \textless object\textgreater~appear in the image? \\
    Count all the instances whose category is the same as \textless object\textgreater~present in the image.
\end{tcolorbox}

\noindent\textbf{Instruction templates of object detection.}
We list all the instruction templates of object detection task below. \textless category\textgreater~will be replaced by the category name.

\begin{tcolorbox}[title=\small{Instruction templates of object detection\#1}]
    \small
    Locate and mark the positions of all \textless category\textgreater~in the image. \\
    Find all the instances of \textless category\textgreater~in the image and indicate their respective locations. \\
    Spot and record the coordinates of every \textless category\textgreater~present in the image. \\
    Identify the \textless category\textgreater~in the image and provide their precise locations. \\
    Can you determine the positions of all the \textless category\textgreater~in the image and list them? \\
    Pinpoint the \textless category\textgreater~in the image and give me their exact coordinates. \\
    Locate all the \textless category\textgreater~in the image and provide their locations in detail. \\
    Detect and report the locations of all the \textless category\textgreater~present in the image. \\
    Find and list the locations of every \textless category\textgreater~in the image. \\
    Please identify the \textless category\textgreater~in the image and give me their locations. \\
    Provide me with the precise locations of all the \textless category\textgreater~in the image. \\
    Detect and record the positions of the \textless category\textgreater~in the image. \\
    Spot all the instances of \textless category\textgreater~in the image and give me their coordinates. \\
    Detect all the \textless category\textgreater~in the image, and output there location. \\
    There are some \textless category\textgreater~in the image, could you help me to locate them and give me their coordinates. \\
    What are the coordinates of the \textless category\textgreater~in the image. \\
    Give the detail locations of all the \textless category\textgreater~you can find in the image.
\end{tcolorbox}

\begin{tcolorbox}[title=\small{Instruction templates of object detection\#2}]
    \small
    Locate all the items in the picture that share the same category as \textless object\textgreater~and provide their coordinates. \\
    Spot every object that belongs to the same category as \textless object\textgreater~and indicate their positions. \\
    Identify all the objects that fit the same category as \textless object\textgreater~and display their coordinates. \\
    Find all the objects that have a similar classification as \textless object\textgreater~and output their locations. \\
    Locate and report the coordinates of all the objects that share the category with \textless object\textgreater. \\
    Detect all the objects in the image that have the same classification as \textless object\textgreater~and provide their positions. \\
    Spot all the objects that belong to the same category as \textless object\textgreater~and show their coordinates. \\
    Identify every instance that falls under the same category as \textless object\textgreater~and report their locations. \\
    Find and output the coordinates of all the objects that have the same category as \textless object\textgreater. \\
    Locate all the objects in the picture that have a similar classification as \textless object\textgreater~and display their positions. \\
    Detect and report the positions of all the objects that share the category with \textless object\textgreater. \\
    Spot every instance that has a similar classification as \textless object\textgreater~and indicate its coordinates. \\
    Identify all the objects that have the same classification as \textless object\textgreater~and output their positions. \\
    Find all the objects that belong to the same category as \textless object\textgreater~and report their locations. \\
    Locate and output the coordinates of all the items that have a similar category as \textless object\textgreater. \\
    Detect all the instances in the image which have the same category with \textless object\textgreater, and output there location. \\
    Detect and report the locations of all the instances present in the image, these instances should have similar category with \textless object\textgreater. \\
    Given an \textless object\textgreater, please help me to find all the instances with the same category. The output should be the coordinates of detected instances.
\end{tcolorbox}

\noindent\textbf{Instruction templates of multi-choices VQA.}
Our instruction tuning dataset also includes the multi-choices VQA, \emph{e.g.}, A-OKVQA. Therefore, we also construct some instruction templates for this task and list them below. Placeholder \textless options\textgreater~will be replaced by the options.

\begin{tcolorbox}[title=\small{Instruction templates of multi-choices VQA}]
    \small
    Please take a look at the image and select the correct answer for \textless question\textgreater~from the options given below \textbackslash n\textless options\textgreater. \\
    Examine the image and select the best matched answer to the question: \textless question\textgreater~from the options given below\textbackslash n\textless options\textgreater. \\
    There are some options\textbackslash n\textless options\textgreater. I have a question for you: \textless question\textgreater~Can you select the best matched answers from the given options based on the image? \\
    Regarding the image, you need to identify the correct answer to the question \textless question\textgreater~from the given options\textbackslash n\textless options\textgreater. \\
    Analyzing the image, can you identify the best matched answer to \textless question\textgreater~from the given options\textbackslash n\textless options\textgreater. \\
    Looking at the image, can you quickly answer my question: \textless question\textgreater. Some potential answers are given in the following options\textbackslash n\textless options\textgreater. \\
    Referring to the image, please select the answer for this question: \textless question\textgreater~from the options\textbackslash n\textless options\textgreater. \\
    Could you please check the image and select the answer for my question: \textless question\textgreater~from the options\textbackslash n\textless options\textgreater. \\
    Here is an image and a question: \textless question\textgreater~for you. Please select an option that can answer the question from the given options\textbackslash n\textless options\textgreater. \\
    For this image, I want to know which option can answer my question: \textless question\textgreater~correctly. The options are\textbackslash n\textless options\textgreater. \\
    Take a look at the image, can you select the best matched answer to the following question: \textless question\textgreater~from following options\textbackslash n\textless options\textgreater.\\
    Considering these options\textbackslash n\textless options\textgreater. I need a correct selection from these options that can answer this question: \textless question\textgreater~in regards to the image.
\end{tcolorbox}

\end{document}